\documentclass{article}

\PassOptionsToPackage{authoryear, sort}{natbib}
\usepackage[preprint]{style/neurips_2026}
\usepackage[utf8]{inputenc} % allow utf-8 input
\usepackage[T1]{fontenc}    % use 8-bit T1 fonts
\usepackage{hyperref}       % hyperlinks
\usepackage{url}            % simple URL typesetting
\usepackage{booktabs}       % professional-quality tables
\usepackage{amsfonts}       % blackboard math symbols
\usepackage{nicefrac}       % compact symbols for 1/2, etc.
\usepackage{microtype}      % microtypography
\usepackage[table]{xcolor}  % colors
\usepackage{graphicx}
\graphicspath{{fig/}}
\usepackage{enumitem}
\usepackage{algorithm}
\usepackage{algpseudocode}

\usepackage{multirow}
\usepackage{amssymb}
\usepackage{subcaption}
\usepackage{makecell}
\usepackage{aux/macros}
\usepackage{adjustbox}
\usepackage{wrapfig}
\usepackage{fontawesome5}

\usepackage{tcolorbox}
\tcbuselibrary{skins,breakable}
\definecolor{promptpurple}{HTML}{ede7f6}   % light lavender
\definecolor{promptborder}{HTML}{9575cd}   % muted purple

\newtcolorbox{thoughtbox}[1][]{%
  enhanced, breakable,
  colback=promptpurple,
  colframe=promptborder,
  boxrule=0pt, leftrule=2.5pt,
  arc=0pt,
  left=6pt, right=4pt, top=3pt, bottom=3pt,
  before skip=6pt, after skip=6pt,
  fontupper=\footnotesize\itshape,
  #1
}

\newcommand{\badgehref}[3]{%
  \href{#1}{%
    \textcolor{black}{%
      \normalsize\raisebox{-0.08em}{#2}\hspace{0.35em}#3%
    }%
  }%
}

\definecolor{darkblue}{rgb}{0, 0, 0.5}
\hypersetup{colorlinks=true, citecolor=darkblue, linkcolor=darkblue, urlcolor=darkblue}

\title{RevengeBench: Reverse Engineering Code-Space Policies from Behavioral Experiments}

\author{
Babak Rahmani\thanks{Equal contribution. Correspondence to rahmani.b91@gmail.com.} \quad
Sebastian Dziadzio\footnotemark[1] \quad
\textbf{Joschka Strüber\footnotemark[1]} \\
\textbf{Sergio Hernández-Gutiérrez} \quad
\textbf{Matthias Bethge} \\[4pt]
Tübingen AI Center
}

\usepackage{amsmath,amsfonts,bm}

\def\eqref#1{equation~\ref{#1}}
\def\1{\bm{1}}

\DeclareMathAlphabet{\mathsfit}{\encodingdefault}{\sfdefault}{m}{sl}
\SetMathAlphabet{\mathsfit}{bold}{\encodingdefault}{\sfdefault}{bx}{n}

\begin{document}
\maketitle

\vspace{-2em}
\begin{center}
\badgehref{https://revengebench.com/\#leaderboard}{\faGlobe}{Leaderboard}
\hspace{2.5em}
\badgehref{https://github.com/bethgelab/revenge-bench}{\faGithub}{Code}
\hspace{2.5em}
\badgehref{https://huggingface.co/datasets/bethgelab/revengebench}{\faDatabase}{Dataset}
\end{center}
\vspace{1.5em}

\begin{abstract}
For most of scientific history, researchers studying behavior could only infer hidden mechanisms from outward actions—an inverse problem that becomes more tractable when observation is augmented by targeted intervention. We pose a computational analogue: given only behavioral traces of an agent in a game environment, can a learner reconstruct the underlying decision program as executable code, and how much does this reconstruction improve with the ability to design controlled experiments? We introduce \benchname, a benchmark of 75 LLM generated, Elo-calibrated policies across five game environments, drawn from CodeClash tournament trajectories. The learner observes the hidden target policy play against sampled opponents and designs behavioral probes in the form of custom opponent policies that elicit informative behavior. It then submits an executable hypothesis, which is evaluated using continuous action-distance metrics. We further validate that recovered code carries informative signal in downstream player-versus-player tournaments. Across twelve frontier LLMs, recovery quality varies substantially (34–72\% of initial distance closed), with reconstructed policies yielding measurable competitive advantage, particularly for weaker models that otherwise struggle to design effective counter-strategies. Our benchmark positions behavioral recovery of programmatic policies as a tractable inverse problem in code-space, opening a path to opponent modeling, policy interpretability, and the broader question of inferring latent mechanisms from observations.
\end{abstract}

\section{Introduction}

\label{intro}

\begin{figure}[h]
    \centering
    \includegraphics[width=\textwidth]{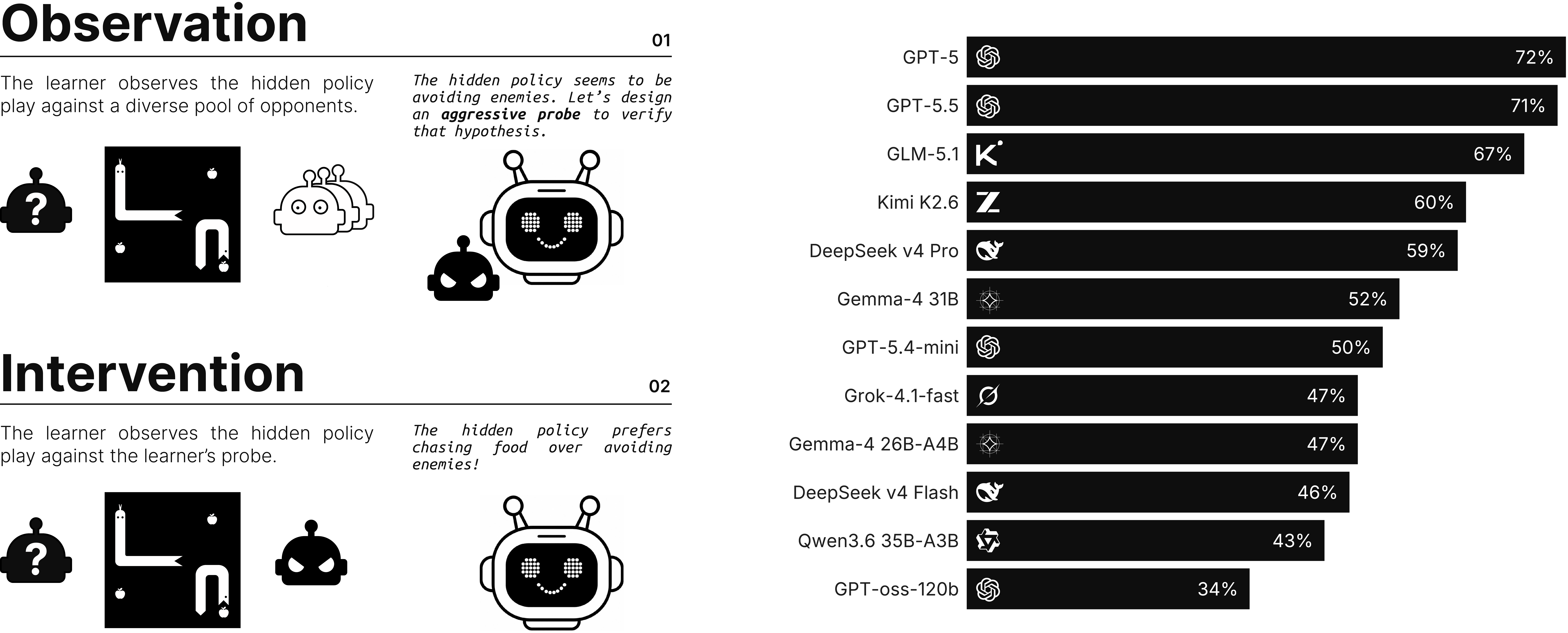}
    \caption{%
        \textbf{Benchmark overview.} \textbf{Left:} the learner alternates between passively observing the hidden policy play against sampled opponents and actively probing it with self-authored opponents. \textbf{Right:} fraction of initial action distance closed ($\uparrow$) for twelve frontier LLMs using \texttt{mini-SWE-agent}.
    }
    \label{fig:teaser}
\end{figure}

Scientific discovery often starts from an asymmetry: effects are visible, but their causes are hidden. Researchers trying to understand animal behavior could not look inside the organisms they studied. Instead, they observed actions, designed experiments, and inferred the latent mechanisms---neural, hormonal, ultimately genetic---that produced what they saw. Such inverse problems become more tractable when passive observation can be paired with active intervention: the same observational evidence can support many candidate mechanisms, but targeted interventions help distinguish among them~\citep{hacking1983representing}. We study a computational analogue: given only behavioral traces of an opaque policy in a game environment, can a learner reconstruct a program that faithfully reproduces the target's behavior? How much does reconstruction improve when the learner can actively probe the target through controlled interactions? Unlike most natural settings, where the true mechanism remains inaccessible, our framework holds the target policy fixed and measurable while varying the learner's epistemic access from passive observation to active probing.

Inverse problems in agent modeling often assume one of two access regimes: a fixed corpus of demonstrations to imitate or invert~\citep{ng2000algorithms, abbeel2004apprenticeship, torabi2018behavioral}, or direct input-output examples with a query interface for arbitrary inputs, as in recent benchmarks for programming by example~\citep{naik2025pbebench, geng2025large, wei2025codearc}. Real scientific inference is usually neither. A biologist, psychologist, or physicist cannot place a system in an arbitrary latent state. They can only perturb it through the channels the environment makes available, and observe the behavior those perturbations elicit. \benchname{} models this intermediate epistemic regime.
The learner cannot query arbitrary target states directly. Instead, it writes \emph{probe opponents} whose interactions with the target induce informative trajectories.
A probe is a hypothesis dressed as an opponent: a policy crafted to drive the target into desired regions of state space. The benchmark thus tests \emph{constrained experimental inference over executable policies}: recovery of programs from behavior, with interventions limited by the dynamics of the environment itself.

Existing accounts of action understanding formalize this problem as recovering a posterior over discrete goals~\citep{baker2009action}, a reward function~\citep{ng2000algorithms, abbeel2004apprenticeship}, or a parametric policy~\citep{torabi2018behavioral}. Each commits the latent mechanism in advance to a particular form, even though strategically important behavior may arise from routines, heuristics, or exceptions rather than from pure utility maximization. A recent line of work argues that behavior can be effectively modeled as executable code: a representation rich enough to encode procedure, memory, and conditionals, and precise enough to be tested at the level of individual actions~\citep{jha2025modeling}. It is already established that LLMs can write executable strategies in the forward direction: recent work synthesises counter-policies in code space~\citep{bachrach2025combining, hennes2026codespaceresponseoraclesgenerating}. We adopt this representation but raise the bar for the learner: rather than weighting an ensemble of candidate programs~\citep{jha2025modeling}, we ask for a single runnable policy, executed on the hidden target's own states across held-out trajectories.

We instantiate this in \benchname{}, a benchmark of 75 executable policies across five game environments from CodeClash~\citep{yang2025codeclash} spanning discrete, grid-based, sequential, multi-unit, and continuous control settings. The policies are written by LLMs for CodeClash competitions. We extract policies from multiple models and pitch them against one another to estimate Elo scores, giving a controlled difficulty axis. In each tournament, the learner observes the target playing against sampled opponents, optionally designs probe opponents, and submits an executable policy.

\begin{figure}[h]
    \centering
    \includegraphics[width=\textwidth]{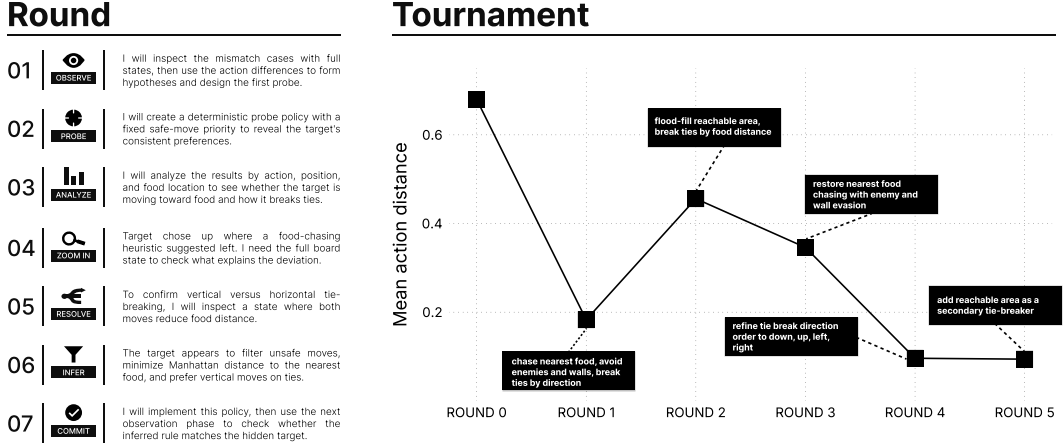}
    \caption{\textbf{Iterative reasoning in BattleSnake. Left:} the reasoning trace of GPT-5 within a single round, all quotes verbatim. \textbf{Right:} GPT-5.4-mini's per-round action distance and submitted strategy summaries: a good initial hypothesis is followed by a regression and a self-correction by round 5.
    }
    \label{fig:traces}
\end{figure}

Figure~\ref{fig:traces} illustrates the inference process of the agent in a single trajectory. The learner moves from mismatch inspection to targeted probing, then revises hypotheses about food pursuit, safety filtering, tie-breaking, and reachable area. The action-distance curve shows that intermediate edits can temporarily worsen recovery before later evidence improves the hypothesis, exposing the iterative nature of behavioral inference.

We evaluate twelve frontier LLM coding agents on \benchname{}. Our primary score is the fraction of action distance closed relative to an initial random policy. Recovery quality varies substantially: the best model closes 72\% of initial behavioral distance, the weakest only 34\%. Active probing helps some models and does not hurt others, indicating that intervention yields benefits when the learner is strong enough to use it. As a secondary check, we test whether behavioral fidelity carries strategic value: we give the recovered program to the same LLM and ask it to write a counter-policy against the target. Counter-policies designed from recovered code have measurable competitive advantage, especially for weaker models that otherwise fail to design effective counters, suggesting it extracts structure that models cannot synthesise from scratch.

Our contributions are threefold. (1) We introduce \benchname{}, consisting of 75 hidden target policies and an evaluation pipeline. (2) We formalise a constrained-intervention recovery protocol in which learners infer executable policies from behavior, with intervention restricted to designing opponent policies that interact with the target through gameplay. (3) We evaluate twelve frontier LLM coding agents, showing that recovery closes 34--72\% of random policy behavioral distance and that recovered code yields counter-policies that are effective against the target. 
The setting matters in two ways. Practically, as LLM agents are increasingly deployed in multi-agent workflows~\citep{openclaw2026}, the ability to anticipate an opaque agent's behavior from observation has direct value. Methodologically, our benchmark opens a path towards understanding when behavioral observation is sufficient for LLM agents to recover a hidden mechanism, and where intervention is required.
\section{Problem Setting}
\label{setting}

Our benchmark operationalizes the inverse task introduced in Section~\ref{intro}: a learner observes a hidden policy playing a game and must submit a runnable program that matches its behavior. Because the hidden policies are themselves executable, hypotheses can be scored mechanically against ground truth, a property that behavioral inverse problems generally lack.

\textbf{Setting.}
Let $\mathcal{S}$ be a state space, $\mathcal{A}$ an action space, and $P(s_{t+1} \mid s_t, a_t)$ the environment dynamics. A \emph{target} is an unknown, fixed policy $\pi^* : \mathcal{S} \to \mathcal{A}$, implemented as an executable strategy. The learner has no access to $\pi^*$ or its internal representation. Interaction occurs through an opponent pool $\mathcal{O}$: for any opponent $o \in \mathcal{O}$, rolling out $\pi^*$ against $o$ yields a trajectory
\[
    \tau = (s_0, a_0, s_1, a_1, \dots, s_T), \quad a_t = \pi^*(s_t).
\]
The learner observes a collection of such trajectories and may optionally generate additional trajectories through controlled probing interactions.

\begin{figure}[t]
\centering
% ── Action structure table (full width) ──
\newcommand{\gameheader}[3][]{%
  \shortstack[c]{%
    {\scriptsize\textsc{#3}}\\[3pt]%
    \includegraphics[#1, width=2cm, height=2cm]{#2}}}
\small
\captionof{table}{\textbf{Action structure.} \textit{Action} is what $\pi^*(s_t)$ returns; $d_g$ is the per-state action distance in Eq.~\ref{eq:mean-action-distance}.}
\setlength{\tabcolsep}{2pt}
\renewcommand{\arraystretch}{1.3}
\begin{tabular}{@{}l
  >{\centering\arraybackslash}m{2.5cm}
  >{\centering\arraybackslash}m{2.5cm}
  >{\centering\arraybackslash}m{2.5cm}
  >{\centering\arraybackslash}m{2.5cm}
  >{\centering\arraybackslash}m{2.5cm}@{}}
    &
    \gameheader{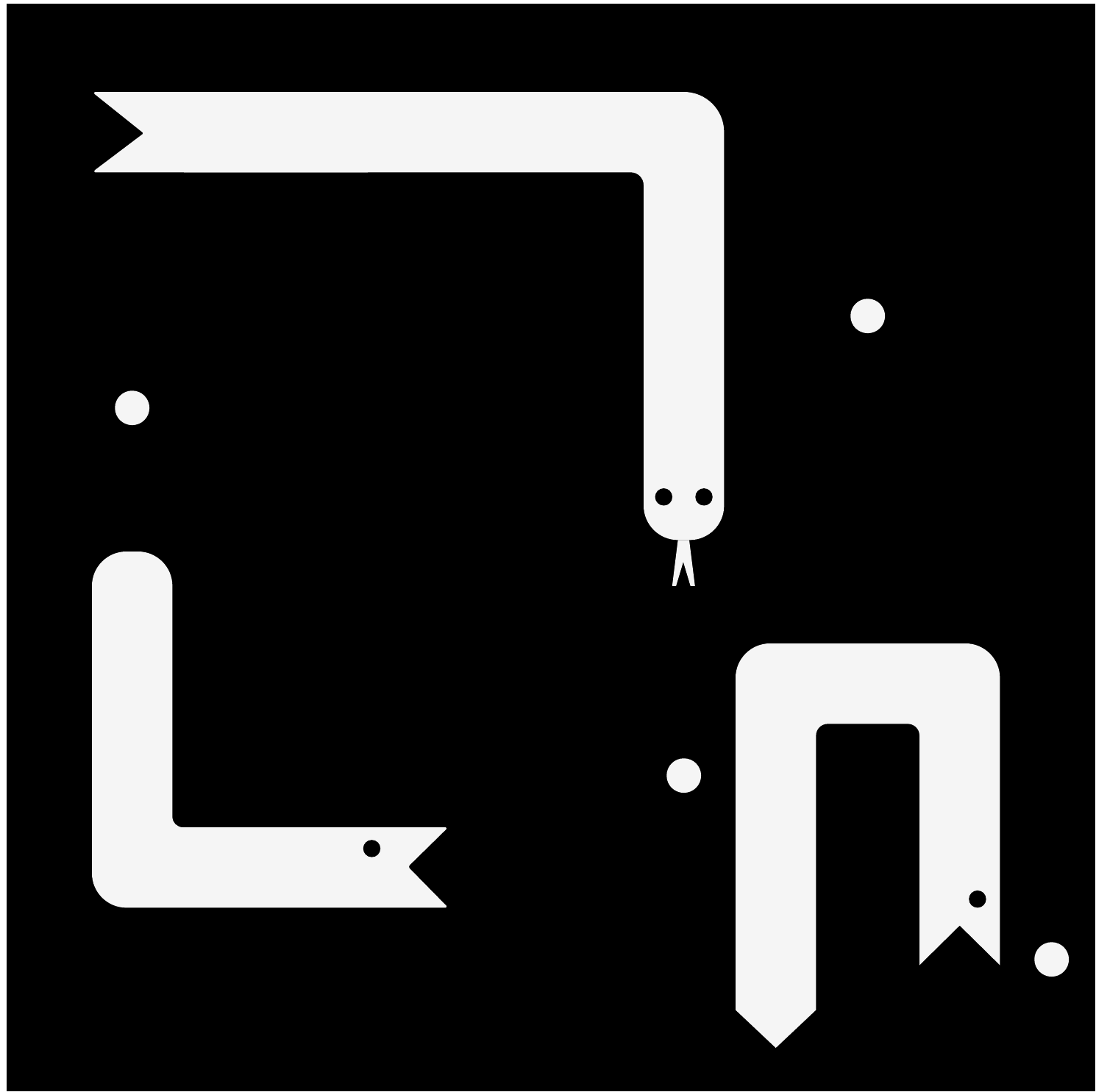}{BattleSnake} &
    \gameheader{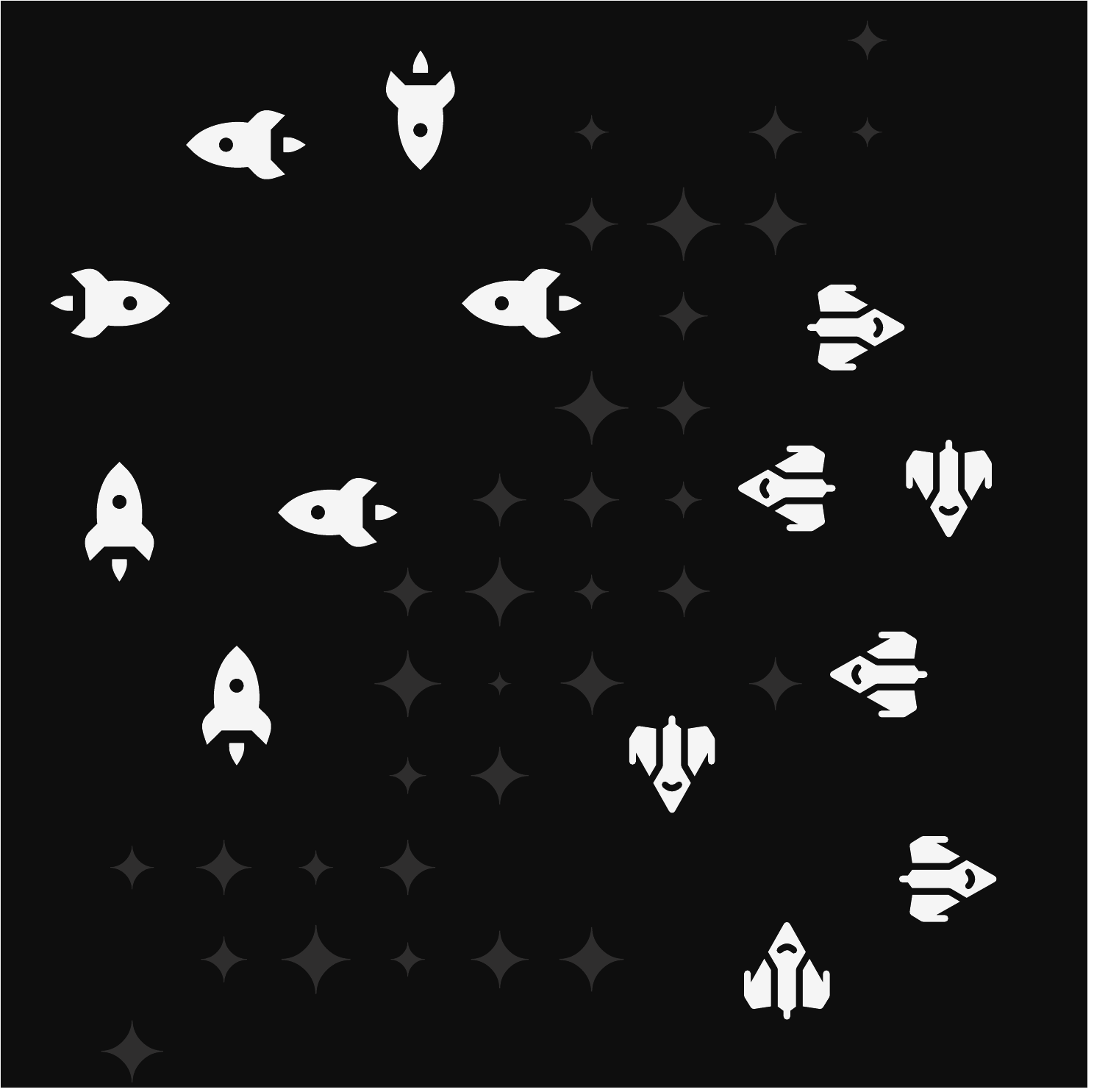}{Halite} &
    \gameheader{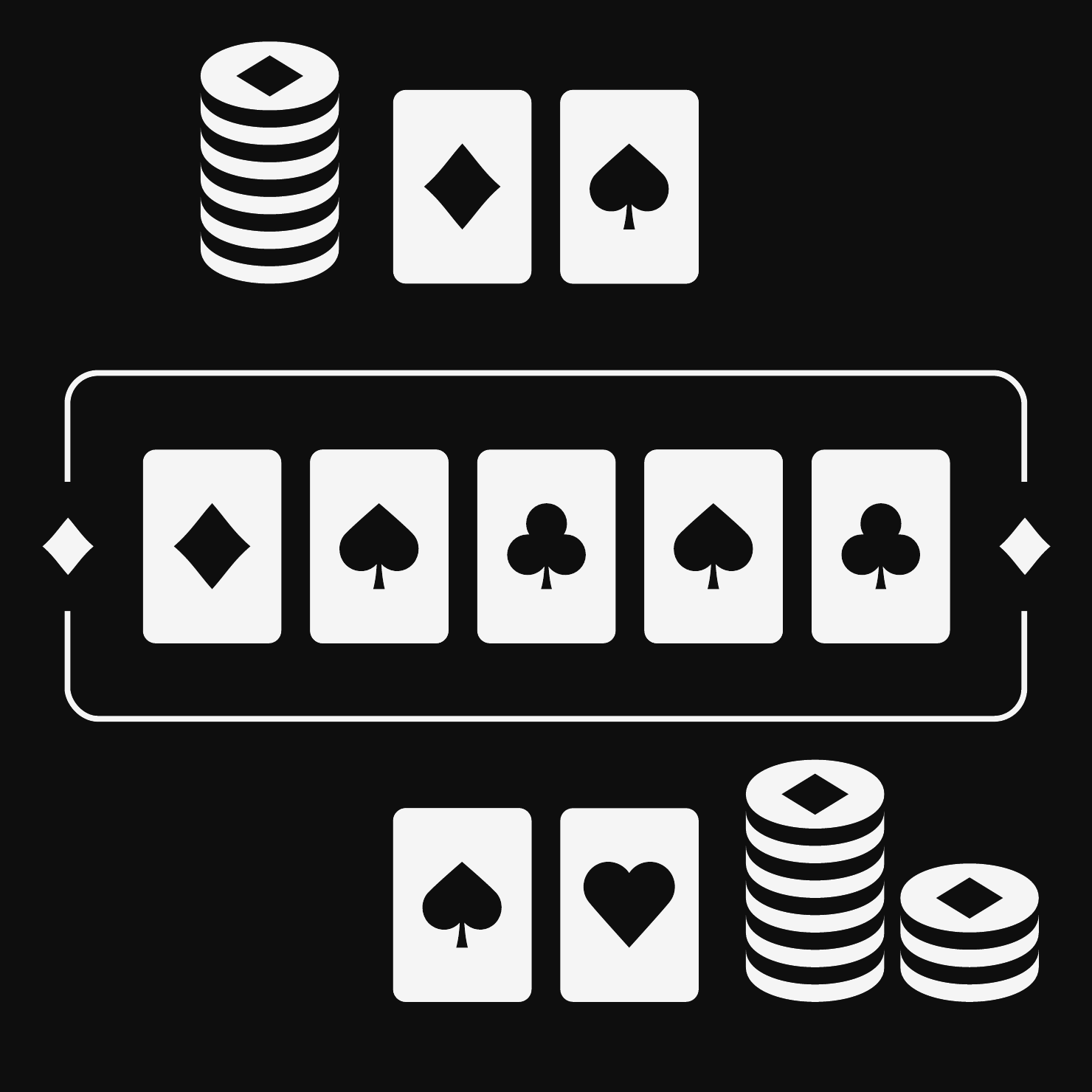}{Poker} &
    \gameheader{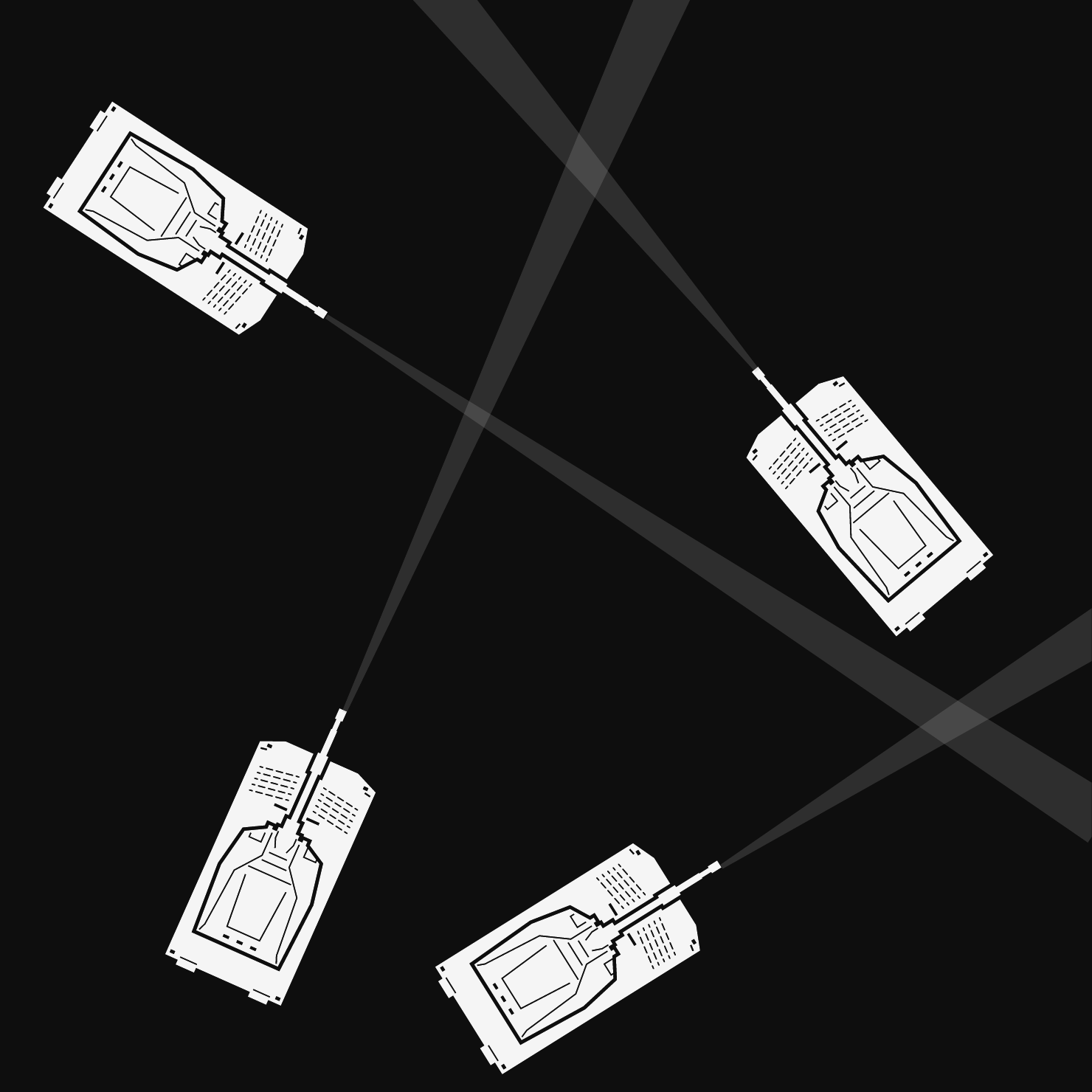}{RoboCode} &
    \gameheader{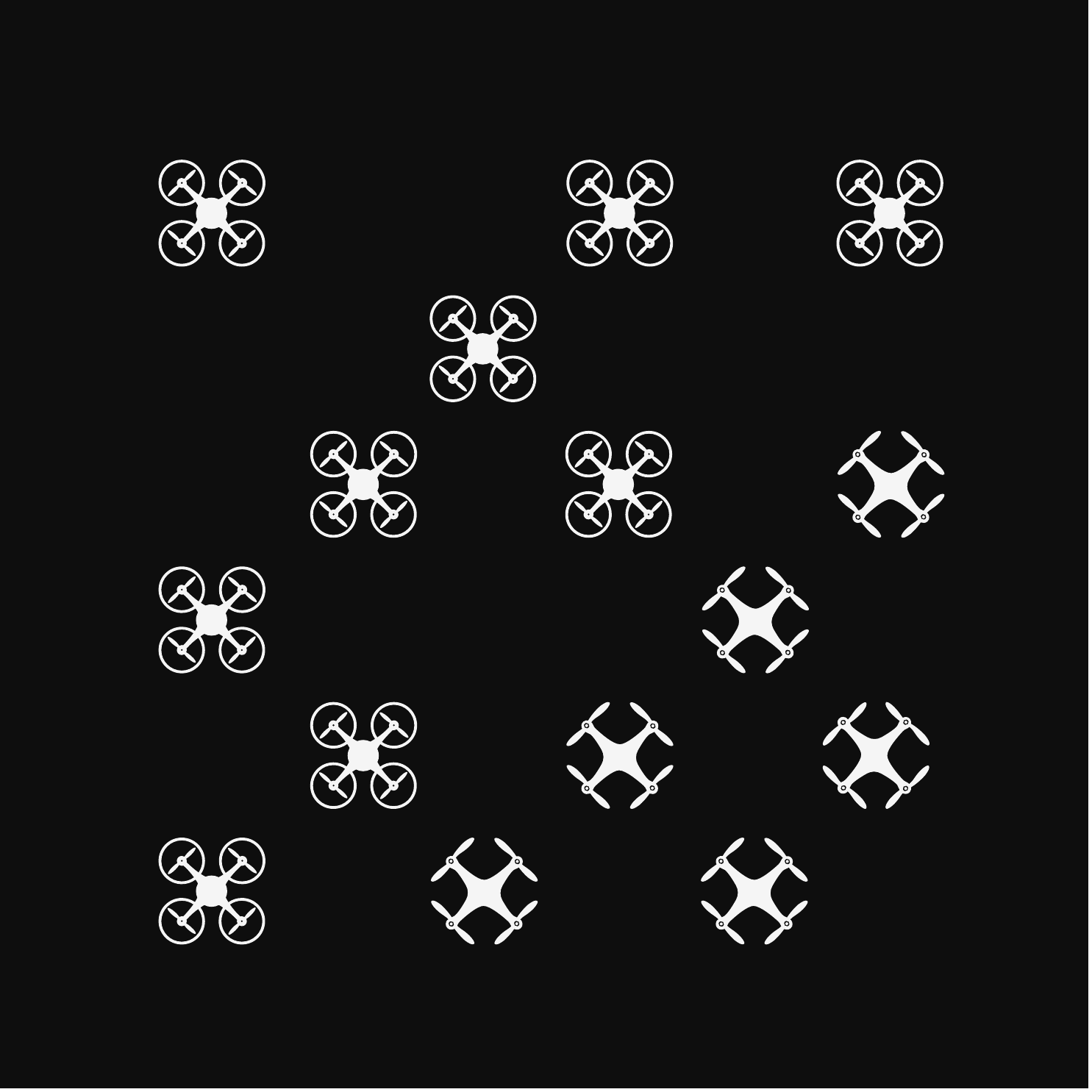}{RobotRumble} \\[4pt]
    \midrule
    \rowcolor{gray!8}
    \rule{0pt}{2.4ex}Action &
    1 of 4 directions &
    per-cell move list &
    fold, check, call, raise($r$) &
    $\mathbb{R}^5$ (vel., turns, fire) &
    per-unit (type, dir.) \\
    \midrule
    $d_g$ &
    $\mathbf{1}[a_1{\neq}a_2]$ &
    fraction mismatched cells &
    log-normalized raise; fold binary &
    mean normalized component difference &
    0/0.5/1 per unit; averaged \\
\end{tabular}
\label{tab:per-game-distance}
\end{figure}

\textbf{Environment and arenas.}
We instantiate the task in five code-based arenas drawn from the CodeClash benchmark \citep{yang2025codeclash}: BattleSnake~\citep{chung2020battlesnake}, Halite~\citep{truell2016halite}, Poker~\citep{kumar2025huskybench}, RoboCode~\citep{hartness2004robocode}, and RobotRumble~\citep{outkine2020robotrumble}.\footnote{CodeClash also includes CoreWar, which we exclude because its action representation does not admit the action-distance comparison required by our evaluation.} 
The five arenas span four programming languages (Python, C, Java, JavaScript) and diverse game mechanics: full versus partial observability, simultaneous versus sequential play, discrete versus continuous control, and action spaces ranging from a single discrete move to structured outputs over hundreds of board locations. Together, these arenas cover a wide range of policy complexity: the most elaborate policies reach roughly 500 lines of code and implement heuristics over evolving game state and opponent behavior (see Tables~\ref{tab:per-game-distance} and~\ref{tab:data-statistics}).

Our framework reuses the CodeClash codebase, which exposes a sandboxed Linux environment in a Docker container. The learner agent acts through the \texttt{mini-SWE-agent} scaffold \citep{yang2024sweagent} by issuing bash commands to read and write files, run scripts, and submit probes or candidate policies for scoring. We deliberately use a minimal harness; the trade-off is discussed in Sec.~\ref{sec:harness}.

\textbf{Policy pool.}
We construct a pool of 40 policies for each arena using CodeClash tournament artifacts. We start from about 15,000 policy files across all arenas and filter them in several stages. First, we apply structural filters to eliminate policies that violate basic constraints. Next, we cluster the remaining files to remove near-duplicates and candidates that do not meet a threshold for semantic complexity. Finally, we verify that each remaining policy is runnable and supports self-play in the corresponding arena container. This process yields 40 validated policies per arena.

For each arena, the 15 highest-Elo policies among these 40 are designated as hidden targets (see Table~\ref{tab:data-statistics}). For each target, the other 39 policies form the opponent pool, from which 20 are sampled each tournament round. Restricting targets to the top tier excludes degenerate entries (e.g., always folding pre-flop in Poker). See Appendix~\ref{app:strategy-pool} for the full construction pipeline.

\textbf{Starter policy.}
Each arena has a simplistic but functional \emph{starter policy} $\hat{\pi}_0$. The learner's hypothesis and probes start from this policy, and our main metric measures improvement relative to it.

\begin{wraptable}{r}{0.6\linewidth}
\vspace{-13pt}
\centering
\small
\caption{\textbf{Target policy statistics.} 75 CodeClash targets (top-15 Elo per game): code length in non-empty LoC and Elo ranges.}
\setlength{\tabcolsep}{5pt}
\begin{tabular}{l l r r r l}
\toprule
\multirow{2}{*}{Game} & \multirow{2}{*}{Language}
  & \multicolumn{3}{c}{\#LoC}
  & \multirow{2}{*}{Elo Range} \\
& & Min & Max & Med. & \\
\midrule
BattleSnake & Python & 139 & 443 & 234 & 1509--1599 \\
Halite & C & 49 & 195 & 86 & 1505--1652 \\
Poker & Python & 85 & 359 & 227 & 1519--1642 \\
RoboCode & Java & 49 & 442 & 125 & 1505--1584 \\
RobotRumble & JS & 29 & 245 & 74 & 1515--1691 \\
\bottomrule
\end{tabular}
\label{tab:data-statistics}
\end{wraptable}

\textbf{Action distance and recovery metric.}
We score recovery using a per-game action distance $d_g: \mathcal{A}_g \times \mathcal{A}_g \to [0, 1]$. Each distance is symmetric, satisfies $d_g(a, a) = 0$, and equals $1$ when either action is missing or invalid. Game-specific distances are aggregated over action components (e.g., Halite cells, RobotRumble units) to handle variable-size action outputs. A graded distance is essential: some policies contain quantitative heuristics (raise sizes, turn rates, per-cell move distributions) for which ``right behavior, wrong magnitude'' is different from ``qualitatively wrong.'' For example, in Poker, betting 90 chips instead of 100 is much closer to the target behavior than folding, even though both would be incorrect under exact-match accuracy. Table~\ref{tab:per-game-distance} summarizes the arena-specific definitions of $d_g$, with full definitions in Appendix~\ref{app:distances}.
The mean per-step action distance on a trajectory $\tau$ is
\begin{equation}
    \label{eq:mean-action-distance}
    D(\hat{\pi}, \pi^*; \tau) \;=\; \frac{1}{T}\sum_{t=1}^{T} d_g\!\left(\hat{\pi}(s_t),\, \pi^*(s_t)\right),
\end{equation}
evaluated on held-out trajectories generated by $\pi^*$ against opponents from $\mathcal{O}$. We report two summaries: $D_{\min}$ ($\downarrow$), the lowest mean action distance achieved across rounds, and the distance reduction ($\uparrow$) $\Delta = 1 - D_{\min}/D_0$, where $D_0$ is the mean action distance of the starter policy $\hat{\pi}_0$. $\Delta$ normalizes for differences in baseline difficulty across targets and arenas, and is used in Fig.~\ref{fig:teaser}.

\textbf{Passive and active access.}
\label{sec:access-regimes}
In the \emph{passive} regime, the learner observes a fixed collection of trajectories generated by $\pi^*$ against opponents sampled from $\mathcal{O}$, with fresh simulations added each round. In the \emph{active} regime, the learner writes up to $B$ probe opponents per round to elicit targeted behavior, for example forcing states where the current hypothesis $\hat{\pi}$ and $\pi^*$ are likely to disagree.

\textbf{Inference as a closed loop.}
The benchmark induces a closed-loop inference problem. Each round, the learner receives fresh trajectories, evaluates its current hypothesis against them, and refines its code. In the active regime, it may interleave probe-writing with refinement so that intervention results feed back into the same round. A more detailed description of the loop and the full version of the algorithm can be found in Appendix~\ref{app:algorithm}.
\section{Experimental Setup}
\label{sec:setup}

We evaluate frontier language models on \benchname{} across all five game environments (BattleSnake, Halite, Poker, RoboCode, and RobotRumble) with 15 targets per game, for a total of 75 targets. Unless otherwise stated, each run consists of $R{=}5$ refinement rounds with 30 edit steps per round. In each round, the learner observes 20 opponent traces and may optionally use up to $B{=}5$ probes.

\textbf{Agent.}
\label{sec:agent}
We use \texttt{mini-SWE-agent}~\citep{yang2024sweagent}, a minimal scaffold that operates inside a sandboxed Linux environment in a Docker container. It has a single tool, a bash terminal, through which it reads and writes files, runs evaluation scripts, and submits candidate policies. This minimal interface reduces scaffold-specific advantages, ensuring comparisons primarily reflect model capability rather than agent engineering.
To compare agent scaffolds, we also evaluate Codex CLI~\citep{openai2025codex} with GPT-5.5~\citep{openai2025gpt55} (high and low reasoning effort) on the same task suite (Section~\ref{sec:analysis}).

\textbf{Models.} \label{sec:models} We evaluate twelve frontier models as learner agents: GPT-5.5, GPT-5~\citep{singh2025openai}, GPT-5.4-mini, GPT-oss-120b, GLM-5.1, Gemma~4~31B, DeepSeek~v4~Pro, DeepSeek~v4~Flash, Gemma~4~26B-A4B, Kimi~K2.6~\citep{team2025kimi}, and Qwen3.6~35B-A3B. Each model is run once per target, for a total of 75 runs per model. Gemma~4~31B is additionally run with five seeds per target for the robustness analysis (Section~\ref{sec:analysis}). For the method-comparison and baseline experiments, we report results on GPT-5, GPT-5.4-mini, Gemma~4~31B, and DeepSeek~v4~Pro. Closed-source models are accessed through their first-party APIs; open-weight models are queried through OpenRouter.

\section{Results}
\label{sec:results-main}

\subsection{Strategy recovery}
\label{subsec:strategy-recovery} 

Figure~\ref{fig:teaser} shows the distance reduction~$\Delta$~per model using the \texttt{mini-SWE-agent} scaffold.
Recovery ranges from 71.9\% (GPT-5) to 33.8\% (GPT-oss-120b), with the top four models clustering above 68\%, a mid-tier (Kimi, DeepSeek~v4~Pro, Gemma-4~31B) recovering 52--60\%, and the remaining models falling between 42--50\%.
All models substantially outperform the pool-mean baseline (Appendix~\ref{app:baseline-pool-mean}), confirming that iterative refinement enables genuine strategy inference rather than lucky pool matching.

\begin{figure}[t]
    \centering
    \begin{minipage}[t]{0.49\linewidth}
        \centering
        \includegraphics[width=\linewidth]{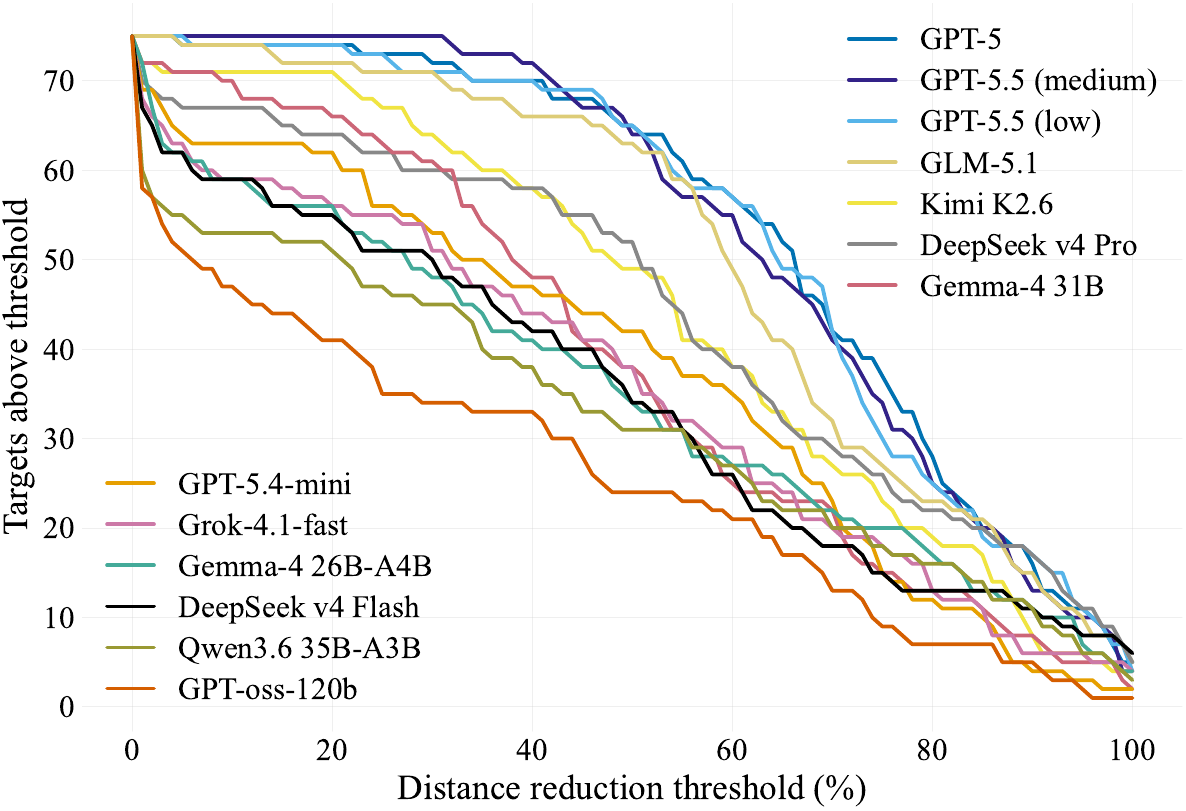}
    \end{minipage}%
    \hfill
    \begin{minipage}[t]{0.49\linewidth}
        \centering
        \includegraphics[width=\linewidth]{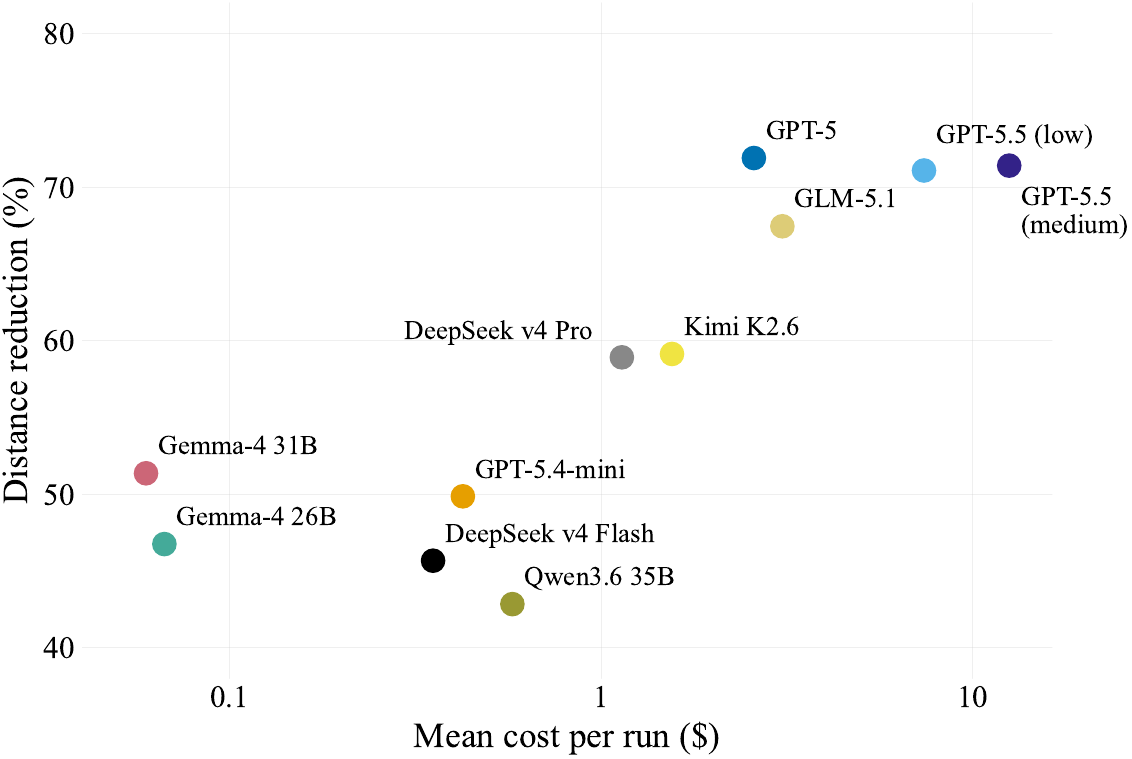}
    \end{minipage}
    \par\vspace{1em}
    \begin{minipage}[t]{\linewidth}
        \centering
        \includegraphics[width=\linewidth]{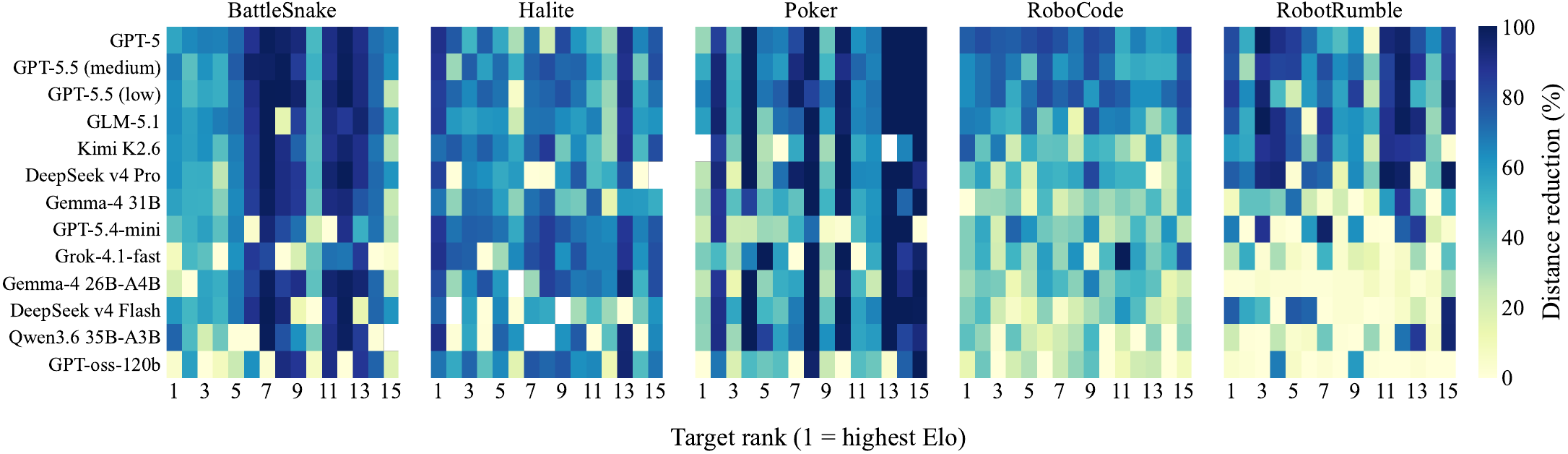}
    \end{minipage}
    \caption{%
        \textbf{Strategy recovery performance.}
        \textbf{Top left:} Cumulative distribution of distance reduction across all 75 targets.
        \textbf{Top right:} Cost--quality trade-off: mean API cost per run vs.\ distance reduction (\%).
        \textbf{Bottom:} Per-model, per-game mean distance reduction (\%).
    }
    \label{fig:recovery-threshold}
\end{figure}

Recovery is widespread but not uniform (Figure~\ref{fig:recovery-threshold}). The strongest models 
fully close the behavioral gap ($\Delta{\geq}99\%$) on a subset of targets (up to 7 of 75 for 
GPT-5.5-low). The cost--quality scatter (top right) reveals three distinct clusters: expensive 
closed models (GPT-5.5 low/medium, \$7--13 per run), a mid-cost tier (GLM-5.1, GPT-5, Kimi, 
DeepSeek~Pro at \$1--3), and a low-cost tier below \$1---where Gemma-4~31B achieves 51\% recovery 
at \$0.06 per run. No meaningful correlation emerges between target playing strength and 
reverse-engineering difficulty (Spearman $\rho{=}0.10$).

Performance varies across arenas (Figure~\ref{fig:recovery-threshold}, bottom). Poker is the 
easiest environment (GPT-5.5-low reaches 82\%; most models exceed 60\%), while RobotRumble 
exhibits the widest model spread (5--72\%). BattleSnake and RoboCode reward the strongest 
models while weaker ones lag considerably. Halite reshuffles the aggregate ranking most: 
mid-tier models such as Gemma-4~26B-A4B (67\%) and GPT-5.4-mini (73\%) match or exceed 
GPT-5/5.5 (61--66\%). Appendix~\ref{app:per-game-recovery} provides the full per-game breakdown.

\subsection{Observation and intervention}
\label{subsec:observation-intervention}

Beyond final recovery, we evaluate how agent's observation mode and inference procedure affect performance. We compare the \textit{Active} and \textit{Passive} regimes from Section~\ref{sec:access-regimes} with two additional baselines: \textit{Description}, which replaces raw traces with natural-language behavioral summaries from an LLM summariser~\citep{hennes2026codespaceresponseoraclesgenerating}, and \textit{Bayesian Program Inference} (BPI), which replaces iterative refinement with best-of-$N$ hypothesis generation followed by Bayesian-style selection~\citep{jha2025modeling}. All modes except BPI use the same iterative code-refinement loop. Implementation details are in Appendix~\ref{app:baseline}.

\begin{table}[!t]
    \centering
    \small
    \caption{\textbf{Method comparison.} Mean$\pm$std best action distance ($\downarrow$) across targets per game. \emph{Pool Mean}: expected distance from sampling $N{=}5$ random pool strategies with no learning; \emph{Probe}: full active method; \emph{No Probe}: passive observation only; \emph{Desc-Only}: natural-language summary input replacing raw traces~\citep{hennes2026codespaceresponseoraclesgenerating}; \emph{BPI}: Bayesian best-of-$N$ ($N{=}5$ to equalise compute budget) from a single observation round~\citep{jha2025modeling}. Cells marked ``--'' indicate invalid code generation by the model. \colorbox{blue!8}{Shaded} rows denote the best method per model.}
    \label{tab:method-comparison}
    \setlength{\tabcolsep}{3pt}
    \renewcommand{\arraystretch}{1.15}
    \begin{tabular}{@{}ll ccccc c@{}}
        \toprule
        Model & Method
          & BattleSnake & Halite & Poker & RoboCode & RobotRumble & Mean \\
        \midrule
        \multicolumn{2}{@{}l}{\textit{Pool Mean (no learning)}}
          & 0.35{\scriptsize$\pm$0.11} & 0.57{\scriptsize$\pm$0.06} & 0.24{\scriptsize$\pm$0.03} & 0.10{\scriptsize$\pm$0.02} & 0.51{\scriptsize$\pm$0.06} & 0.35 \\
        \midrule
        \multirow{4}{*}{\rotatebox[origin=c]{90}{\scriptsize GPT-5}}
          & \cellcolor{blue!8} Probe (Active)   & \cellcolor{blue!8}\textbf{0.15}{\scriptsize$\pm$0.11} & \cellcolor{blue!8}0.31{\scriptsize$\pm$0.17} & \cellcolor{blue!8}\textbf{0.05}{\scriptsize$\pm$0.05} & \cellcolor{blue!8}\textbf{0.08}{\scriptsize$\pm$0.03} & \cellcolor{blue!8}0.14{\scriptsize$\pm$0.11} & \cellcolor{blue!8}\textbf{0.15} \\
          & No Probe         & 0.17{\scriptsize$\pm$0.14} & 0.37{\scriptsize$\pm$0.21} & 0.05{\scriptsize$\pm$0.05} & 0.08{\scriptsize$\pm$0.02} & \textbf{0.12}{\scriptsize$\pm$0.08} & 0.16 \\
          & Desc-Only        & 0.25{\scriptsize$\pm$0.07} & \textbf{0.20}{\scriptsize$\pm$0.08} & 0.09{\scriptsize$\pm$0.07} & 0.16{\scriptsize$\pm$0.04} & 0.16{\scriptsize$\pm$0.11} & 0.17 \\
          & BPI              & 0.27{\scriptsize$\pm$0.11} & 0.32{\scriptsize$\pm$0.22} & 0.15{\scriptsize$\pm$0.14} & 0.17{\scriptsize$\pm$0.06} & 0.22{\scriptsize$\pm$0.22} & 0.23 \\
        \midrule
        \multirow{4}{*}{\rotatebox[origin=c]{90}{\scriptsize GPT-5.4-mini}}
          & Probe (Active)   & 0.32{\scriptsize$\pm$0.19} & 0.21{\scriptsize$\pm$0.07} & \textbf{0.08}{\scriptsize$\pm$0.06} & \textbf{0.17}{\scriptsize$\pm$0.05} & 0.38{\scriptsize$\pm$0.30} & 0.23 \\
          & \cellcolor{blue!8} No Probe         & \cellcolor{blue!8}\textbf{0.21}{\scriptsize$\pm$0.14} & \cellcolor{blue!8}0.24{\scriptsize$\pm$0.11} & \cellcolor{blue!8}0.10{\scriptsize$\pm$0.08} & \cellcolor{blue!8}0.17{\scriptsize$\pm$0.04} & \cellcolor{blue!8}0.44{\scriptsize$\pm$0.31} & \cellcolor{blue!8}\textbf{0.23} \\
          & Desc-Only        & 0.42{\scriptsize$\pm$0.18} & \textbf{0.20}{\scriptsize$\pm$0.07} & 0.10{\scriptsize$\pm$0.08} & 0.21{\scriptsize$\pm$0.04} & 0.84{\scriptsize$\pm$0.25} & 0.35 \\
          & BPI              & 0.47{\scriptsize$\pm$0.17} & 0.25{\scriptsize$\pm$0.16} & 0.16{\scriptsize$\pm$0.24} & 0.26{\scriptsize$\pm$0.04} & \textbf{0.25}{\scriptsize$\pm$0.16} & 0.28 \\
        \midrule
        \multirow{3}{*}{\rotatebox[origin=c]{90}{\scriptsize \shortstack{DeepSeek\\v4 Pro}}}
          & \cellcolor{blue!8} Probe (Active)   & \cellcolor{blue!8}\textbf{0.19}{\scriptsize$\pm$0.14} & \cellcolor{blue!8}0.44{\scriptsize$\pm$0.24} & \cellcolor{blue!8}\textbf{0.05}{\scriptsize$\pm$0.06} & \cellcolor{blue!8}\textbf{0.16}{\scriptsize$\pm$0.05} & \cellcolor{blue!8}0.19{\scriptsize$\pm$0.17} & \cellcolor{blue!8}\textbf{0.20} \\
          & No Probe         & 0.27{\scriptsize$\pm$0.21} & 0.52{\scriptsize$\pm$0.26} & 0.12{\scriptsize$\pm$0.24} & 0.18{\scriptsize$\pm$0.05} & 0.20{\scriptsize$\pm$0.16} & 0.26 \\
          & Desc-Only        & 0.28{\scriptsize$\pm$0.09} & \textbf{0.20}{\scriptsize$\pm$0.09} & -- & 0.17{\scriptsize$\pm$0.03} & \textbf{0.14}{\scriptsize$\pm$0.11} & 0.20 \\
        \midrule
        \multirow{3}{*}{\rotatebox[origin=c]{90}{\scriptsize \shortstack{Gemma-4\\31B}}}
          & Probe (Active)   & \textbf{0.21}{\scriptsize$\pm$0.15} & 0.36{\scriptsize$\pm$0.14} & 0.06{\scriptsize$\pm$0.05} & \textbf{0.19}{\scriptsize$\pm$0.04} & 0.33{\scriptsize$\pm$0.12} & 0.23 \\
          & \cellcolor{blue!8} No Probe         & \cellcolor{blue!8}0.24{\scriptsize$\pm$0.19} & \cellcolor{blue!8}0.30{\scriptsize$\pm$0.22} & \cellcolor{blue!8}\textbf{0.05}{\scriptsize$\pm$0.06} & \cellcolor{blue!8}0.20{\scriptsize$\pm$0.04} & \cellcolor{blue!8}\textbf{0.31}{\scriptsize$\pm$0.09} & \cellcolor{blue!8}\textbf{0.22} \\
          & Desc-Only        & 0.42{\scriptsize$\pm$0.26} & \textbf{0.29}{\scriptsize$\pm$0.27} & -- & 0.19{\scriptsize$\pm$0.05} & 0.46{\scriptsize$\pm$0.28} & 0.34 \\
        \bottomrule
    \end{tabular}
\end{table}

\textbf{Active probing yields model-dependent gains.}
Table~\ref{tab:method-comparison} shows that raw behavioral traces are the strongest observation channel. Passive refinement is already competitive: probing improves recovery in 16 of 20 model--game pairs (see Figure~\ref{fig:probe-boost}), but the benefit is concentrated in the stronger models. DeepSeek~v4~Pro gains the most (mean boost $+0.052$, with the largest effects in BattleSnake, Halite and Poker), followed by GPT-5 ($+0.013$, four of five games). GPT-5.4-mini and Gemma~4~31B show near-zero average gains, as positive and negative effects cancel across games. We analyse the mechanisms behind these model-dependent probing gains in Section~\ref{sec:agent-behaviour}.

\textbf{Baselines.} Raw observations are difficult to compress or replace. Description-only 
inputs are sometimes competitive, especially on Halite, but generally lose fine-grained 
state--action evidence. BPI underperforms iterative refinement for GPT-5 and GPT-5.4-mini; 
DeepSeek~v4~Pro and Gemma~4~31B were unable to submit valid code within BPI's single-round 
editing budget.

\subsection{From Inference to Action}
\label{sec:inference-action}

Strategy recovery produces a computational model of the target's decision procedure.
But does possessing such a model confer a practical advantage?
We test this by asking whether an agent armed with the recovered code can \emph{win more often} against the target than one operating blind.

\textbf{Setup.}
We run a player-versus-player (PvP) tournament in which a challenger LLM iteratively writes a counter-strategy against a fixed target over 5 rounds.
The challenger sees one of three levels of opponent information:
(a)~\textbf{blind} (game rules only),
(b)~\textbf{recovered} (the code produced by our inverse pipeline), or
(c)~\textbf{oracle} (the target's true source code).
Five challenger models of varying strength are evaluated across four games (\textsc{BattleSnake}, \textsc{Halite}, \textsc{RoboCode}, \textsc{RobotRumble}) with 5 targets per game (20 targets total), yielding 540 tournament runs and ${\sim}$81K simulations (full protocol in Appendix~\ref{sec:forward-pvp}).
We measure each challenger's win rate per round and define the \emph{win rate gain} as the win-rate difference between a given condition (oracle or recovered) and blind ($\mathrm{WR}_{\text{condition}} - \mathrm{WR}_{\text{blind}}$).

\begin{figure}[t]
    \centering
    \includegraphics[width=\linewidth]{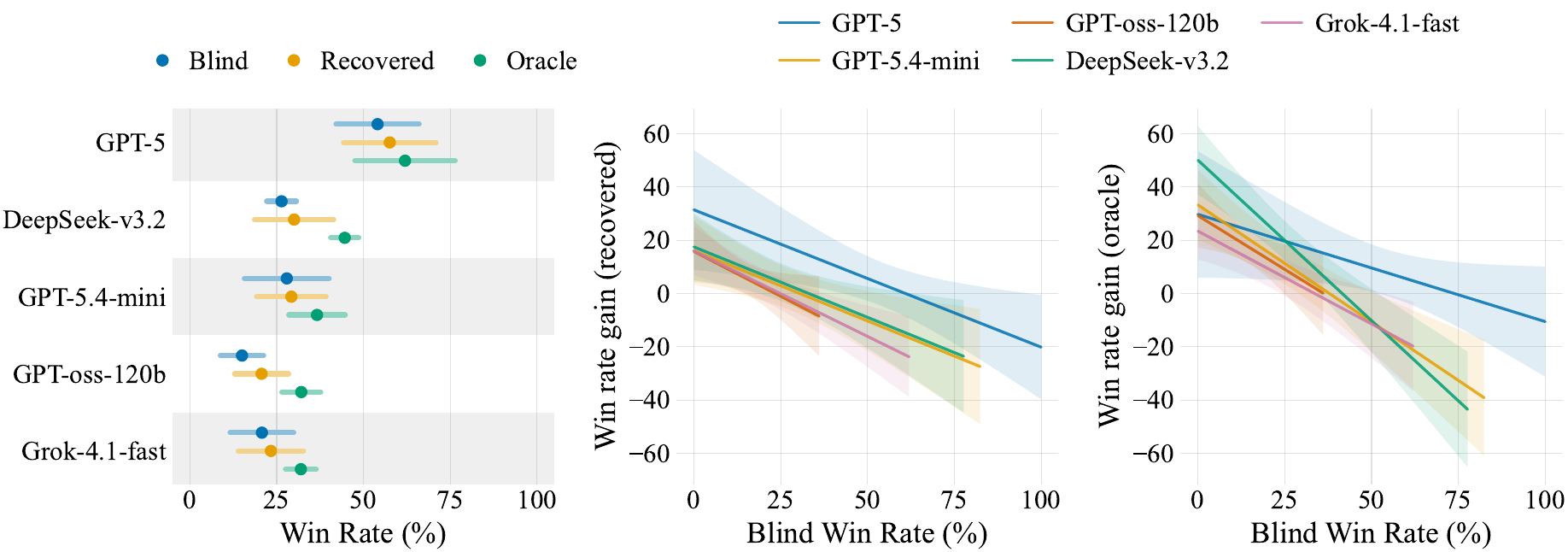}
    \caption{%
        \textbf{PvP tournament}: a challenger agent writes a counter-strategy against a fixed target over 5 rounds, under three levels of opponent information (blind, recovered, oracle).
        Results are averaged over 20 targets (5 per game), 5 rounds, and 3 seeds per model.
        \textbf{Left:} Win rate per challenger model, averaged across games and rounds.
        Horizontal ticks show the cross-game mean.
        The ordering oracle $>$ recovered $>$ blind holds for every model is observed.
        \textbf{Middle, Right:} Each point is one (model, game, target) combination.
        Regression lines with 95\% CI bands show that models with lower blind baselines gain more from opponent intelligence, for both oracle (middle) and recovered (right) conditions.
    }
    \label{fig:pvp-main}
\end{figure}

\textbf{Recovered policy carries useful signal.}
Figure~\ref{fig:pvp-main} (left) shows that across all models the ordering \emph{blind $<$ recovered $<$ oracle} holds consistently: access to the recovered strategy lifts win rates above blind play, and the oracle provides a further ceiling.
The recovered code is imperfect, yet it carries enough actionable signal to produce measurable competitive advantage.

\textbf{Weaker models benefit more.}
The value of opponent intelligence depends on the challenger's baseline ability.
Figure~\ref{fig:pvp-main} (middle, right) shows a significant negative correlation between blind win rate and win rate gain: models that perform poorly without opponent information derive the largest gains from it.
Strong models discover competitive strategies through blind play in later rounds, eroding the initial advantage; weaker models never close this gap, so the win rate gain persists or grows (Appendix~\ref{sec:forward-pvp-h2},~\ref{sec:forward-pvp-h3}).

\section{Ablations and Robustness}
\label{sec:analysis}

\subsection{Full conversation history outperforms per-round reset}
\label{sec:context-management}

Our agent carries its full conversation history across rounds, with older observations compacted via rule-based summaries (Appendix~\ref{app:history-management}). 

\begin{wraptable}{r}{0.46\textwidth}
\vspace{-8.mm}
\centering
\small
\captionof{table}{\textbf{Persistent vs.\ reset}: fraction of initial action distance closed (\%).}
\label{tab:reset-vs-persistent}
% \vspace{2mm}
\begin{tabular}{lrr}
\toprule
\textbf{Model} & \textbf{Pers.} & \textbf{Reset} \\
\midrule
\cellcolor{blue!8} GPT-5 & \cellcolor{blue!8} 71.9 & \cellcolor{blue!8} 67.2 \\
GPT-5.4-mini   & 49.9 & 46.6 \\
Grok-4.1-fast  & 47.2 & 46.1 \\
GPT-oss-120b   & 33.8 & 30.8 \\
\bottomrule
\end{tabular}
\vspace{-7mm}
\end{wraptable}

Clearing the history each round consistently degrades recovery (Table~\ref{tab:reset-vs-persistent}). Both conditions retain a second, agent-driven channel of round-to-round memory~\citep{yang2025codeclash}, so the gap reflects information that persistent context carries beyond the agent's own summaries. Compaction reduces API cost by 11--22\% without measurable performance loss (Appendix~\ref{sec:compaction-ablation}).

\subsection{Scaffold and reasoning effort: which matters more?}
\label{sec:harness}
\begin{table}[t]
    \centering
    \small
    \caption{\textbf{Codex vs.\ mini-SWE-agent.} Mean$\pm$std best action distance ($\downarrow$) across 15 pool targets per game. \emph{Low} / \emph{high} refer to the model's reasoning effort.}
    \label{tab:codex-vs-miniswe}
    \setlength{\tabcolsep}{3pt}
    \renewcommand{\arraystretch}{1.15}
    \begin{tabular}{@{}l ccccc c@{}}
        \toprule
        Method
          & BattleSnake & Halite & Poker & RoboCode & RobotRumble & Mean \\
        \midrule
        GPT-5.5 low (\texttt{mini-SWE-agent})
          & 0.17{\scriptsize$\pm$0.15} & 0.29{\scriptsize$\pm$0.17} & 0.03{\scriptsize$\pm$0.03} & 0.09{\scriptsize$\pm$0.02} & 0.17{\scriptsize$\pm$0.12} & 0.15 \\
        GPT-5.5 low (Codex)
          & 0.14{\scriptsize$\pm$0.13} & 0.26{\scriptsize$\pm$0.13} & 0.03{\scriptsize$\pm$0.04} & 0.11{\scriptsize$\pm$0.03} & 0.09{\scriptsize$\pm$0.09} & 0.12 \\

        \cellcolor{blue!8} GPT-5.5 high (Codex)
          & \cellcolor{blue!8}\textbf{0.13}{\scriptsize$\pm$0.12} & \cellcolor{blue!8}\textbf{0.14}{\scriptsize$\pm$0.13} & \cellcolor{blue!8}\textbf{0.02}{\scriptsize$\pm$0.03} & \cellcolor{blue!8}\textbf{0.07}{\scriptsize$\pm$0.03} & \cellcolor{blue!8}\textbf{0.05}{\scriptsize$\pm$0.07} & \cellcolor{blue!8}\textbf{0.08} \\
        \bottomrule
    \end{tabular}
\end{table}

One concern when evaluating LLMs as agents is the influence of the scaffold and reasoning effort. We address this with an ablation that fixes the model to GPT-5.5 and varies the scaffold (\texttt{mini-SWE-agent} vs.\ Codex CLI) and the reasoning effort (low vs.\ high), evaluating on the same 15-target pool. Table~\ref{tab:codex-vs-miniswe} summarises mean best action distances per game.

\textbf{Both factors matter comparably.} Switching from \texttt{mini-SWE-agent} to Codex at low effort lowers the cross-game mean from $0.15$ to $0.12$. Raising effort from low to high within Codex lowers it further to $0.08$. The two effects are similar and stack additively for a total relative reduction of 47\% from mini-SWE-low to Codex-high.

\textbf{mini-SWE-agent supports cross-model comparison.} Because the Codex CLI is restricted to GPT models, we use \texttt{mini-SWE-agent} as the common scaffold and report the Codex result as a per-model upper-bound point for GPT-5.5.

\textbf{Higher effort helps everywhere.} At fixed harness, high effort improves every game (Halite halves: $0.26 \to 0.14$; three games reach $\le 0.07$).

\subsection{Benchmark robustness}
\label{sec:benchmark-robustness}
\begin{table}[t!]
\centering
\caption{\textbf{Per-arena reliability summary:} \emph{Within-model} columns are the Gemma~4~31B 5-run sweep (15~targets~$\times$~5~seeds; target and round-by-round opponent set fixed). \emph{Cross-model} columns aggregate one run per evaluator across 12~models (Gemma~4~31B at its 5-run mean; the other 11 at \texttt{seed}=42). ICC is the Shrout--Fleiss two-way random-effects, single-rater consistency form on the (target~$\times$~run) or (target~$\times$~model) matrix of normalised recovery score; $\bar\rho$ is the mean pairwise Spearman correlation between per-target rankings; the recovery 95\% CI is from a 2{,}000-resample hierarchical bootstrap.}
\label{tab:reliability-summary}
\begin{tabular}{lccccc}
\toprule
& \multicolumn{3}{c}{Within-model (Gemma 4 31B, 5 runs)} & \multicolumn{2}{c}{Cross-model (12 evaluators)} \\
\cmidrule(lr){2-4} \cmidrule(lr){5-6}
Arena & ICC & $\bar\rho$ & recovery (95\% CI) & ICC & $\bar\rho$ \\
\midrule
BattleSnake & 0.81 & 0.88 & 66.4\% [54.7, 78.0] & 0.32 & 0.51 \\
Halite      & 0.07 & 0.14 & 63.1\% [55.6, 69.3] & 0.20 & 0.33 \\
Poker  & 0.85 & 0.84 & 68.5\% [55.5, 80.4] & 0.58 & 0.61 \\
RoboCode    & 0.41 & 0.32 & 33.7\% [27.9, 39.8] & 0.07 & 0.09 \\
RobotRumble & 0.19 & 0.23 & 30.5\% [22.3, 38.6] & 0.11 & 0.14 \\
\bottomrule
\end{tabular}
\end{table}
To check the internal validity of \benchname{}'s headline metrics, we run Gemma~4~31B on all 75 targets with five seeds (375 runs) and the remaining 11 evaluators at one seed each. Each repetition holds the target and the round-by-round opponent set fixed; only process-level non-determinism (game-engine RNG, agent shuffle, LLM sampling) varies. We decompose the round-5 mean action distance into three nested noise sources: within-run $\sigma_{\rm sim}$ (opponent identity varies across the 20 sims of a round), across-run $\sigma_{\rm seed}$ (only the run varies), and across-target $\sigma_{\rm target}$ (the signal).

Reliability is arena-dependent (Table~\ref{tab:reliability-summary}). Poker and BattleSnake are signal-dominated within Gemma's 5-run sweep (intra-class consistency ICC of 0.85 and 0.81; cross-run Spearman $\rho \geq 0.84$), and Poker remains the only arena with strong cross-evaluator agreement (ICC 0.58, $\bar\rho = 0.61$). Halite, RoboCode, and RobotRumble have within-Gemma ICC $\leq 0.41$ and cross-evaluator ICC $\leq 0.20$: run-to-run noise is comparable to the across-target signal, and evaluators disagree on per-target difficulty. Single-run scores on Poker and BattleSnake therefore support direct model rankings; on the other three arenas, fine-grained model comparison requires run-aggregated reporting, following recent calls for explicit uncertainty quantification on compact reasoning benchmarks \citep{hochlehnert2025a}. A full noise decomposition is given in Appendix~\ref{app:noise-decomposition}.

\section{Agent Behavior Analysis}
\label{sec:agent-behaviour}

\textbf{Trajectory analysis.}
\label{subsec:trajectories}
Following \citet{yang2026programbench}, we classify each bash command into one of five action types by keyword matching in priority order: \emph{Probe} (reference executable invocation via \texttt{PROBE\_SUBMIT}), \emph{Write} (file creation or editing, e.g.\ heredocs, shell redirects, \texttt{sed~-i}), \emph{Execute} (non-probe program execution such as \texttt{python3}), \emph{Read} (file inspection, directory listing, searches), or \emph{Other}.

\begin{figure}[t!]
    \centering
    \includegraphics[width=\linewidth]{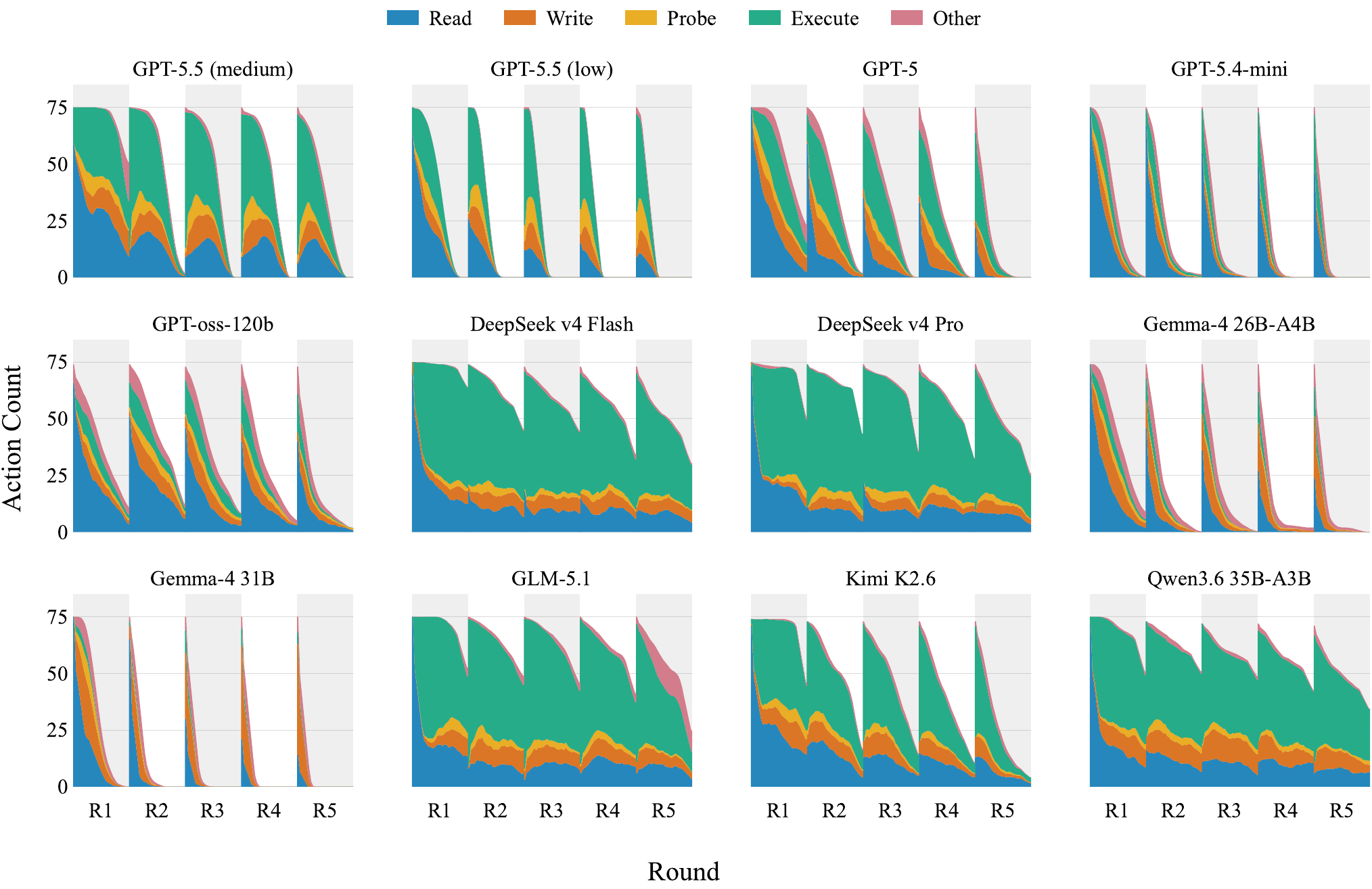}
    \caption{\textbf{Step-level action distribution across rounds} for twelve models over all 75 targets. Cumulative height reflects the number of trajectories still active at each step.}
        \label{fig:action-distribution}
    \vspace{-6mm}
\end{figure}

Turn counts vary substantially across models (Figure~\ref{fig:action-distribution}). Three distinct behavioral clusters emerge. \emph{High-throughput models} (DeepSeek~v4~Flash, DeepSeek~v4~Pro, GLM-5.1, Qwen3.6~35B-A3B) maintain near-constant activity at the 75-target ceiling across all five rounds, dominated by Execute actions with Read as the secondary category; these models iterate rapidly through code-run-inspect cycles and routinely exhaust the 30-step budget per round. \emph{Mid-throughput models} (GPT-5.5-medium, Kimi~K2.6, GPT-5, GPT-5.5-low, GPT-oss-120b) show more varied action profiles and visible round-over-round decay as targets converge or exhaust: GPT-5.5-medium favours Execute and Read, while GPT-oss-120b inverts this pattern. \emph{Low-throughput models} (Gemma-4~26B-A4B, GPT-5.4-mini, Gemma-4~31B) drop sharply within the first few steps of each round. Gemma-4~31B spends the plurality of turns on Write with very little Execute, suggesting it attempts large edits rather than incremental test-fix loops.

\textbf{Why does probing help selectively?}
A probe is an executable experiment the learner must design, implement, and deploy as valid code.

\emph{Reliability} varies: failure rates range from near-zero (GPT-5.5 variants) to the highest values for Grok-4.1-fast and Gemma-4~31B (Figure~\ref{fig:probe-failures}).
\emph{Design} matters more than volume: stronger models craft targeted experiments:
\begin{thoughtbox}
\upshape\textbf{GPT-5} (BattleSnake):\quad\itshape a deterministic, safety-aware policy that avoids walls and occupied cells [\ldots] This should lead to longer survival and more diverse states, giving better insight into the target's decision rules
\end{thoughtbox}
\noindent whereas weaker models default to minimal baselines (e.g.\ GPT-5.4-mini frequently submits a probe that \emph{``always stays still''} or \emph{``does nothing''} to elicit the target's behavior against a static reference).

\textbf{Common failure modes.}
Trajectory inspection reveals three recurring failure patterns that affect all models to varying degrees.

\emph{Step-budget exhaustion.}
High-throughput models frequently hit the 30-step ceiling before submitting (68--74\% of rounds for the high-; 38--39\% for mid-throughput models).
A common trigger is repeated analysis scripts that fail on the same bug:
\begin{thoughtbox}
\upshape\textbf{GPT-oss-120b} (BattleSnake, round~2, step~24 of~30):\quad\itshape Identify mismatched entries where the target's chosen move does NOT maximize distance from opponent heads. Print those states to examine possible patterns\ldots\quad\upshape$\rightarrow$~\texttt{SyntaxError} \emph{(same error repeated at steps 26, 31)}
\end{thoughtbox}
\noindent Low-throughput models (GPT-5.4-mini, Gemma-4~31B, Gemma-4~26B) almost never exhaust their budget ($\leq$\,3\% of rounds), as they issue far fewer commands.

\emph{Regression from best submission.}
Editing a submission can make it worse.
Every model exhibits this: 16--55\% of targets end with a final distance worse than the best achieved in an earlier round.
Stronger models regress least, while weaker models regress in up to half their targets.
\begin{thoughtbox}
\upshape\textbf{GPT-5} (BattleSnake, round~4):\quad\itshape The latest round shows a significant regression, suggesting my last change (vertical-only greedy and down>up>right>left fallback) diverges from the target's true behavior.
\end{thoughtbox}

\emph{Invalid submissions.}
Some models submit code that crashes at evaluation time.
Grok-4.1-fast is the most affected, producing at least one evaluation failure in 44\% of its runs, spread across all five games.
GPT-oss-120b fails in 23\% of runs, with errors in all games.
The remaining models cluster at 0--8\%, with failures concentrated in Halite.
\begin{thoughtbox}
\upshape\textbf{GPT-5.4-mini} (HuskyBench, round~5, after \texttt{nan} distance):\quad\itshape The latest bot caused an evaluation failure with NaN distance, so I need to inspect the current \texttt{client/player.py} for invalid expressions or bad return handling.
\end{thoughtbox}
\section{Related Work}
\label{sec:related_works}

We focus on work closest to \benchname{}: recovering executable policies from behavioral evidence in strategic environments; broader links to code-space policy generation, agent benchmarks, theory of mind, and opponent modelling are discussed in Appendix~\ref{app:extended-related-work}.

\paragraph{Opponent modelling and code-space games.}
Opponent modelling infers an opponent's policy from observed actions to improve decision-making~\citep{nashed2022survey}, from opponent-aware learning and shaping~\citep{foerster2017learning, lu2022model, khan2023scaling, duque2024advantage} to in-context opponent search~\citep{jing2024opponent} and LLM-based opponent models~\citep{yu2025llm, liu2026scaling}. Strategic-agent benchmarks instead evaluate how well LLMs play repeated, game-theoretic, or multi-agent interactions~\citep{akata2025playing, duan2024gtbench, costarelli2024gamebench, piatti2024cooperate}. Code-space game work shows that LLMs can generate executable policies through self-play or response-oracle methods~\citep{bachrach2025combining, hennes2026codespaceresponseoraclesgenerating}. \benchname{} studies the inverse problem: recovering another agent's policy as executable code from behavior, then testing whether that recovered code improves downstream play in Section~\ref{sec:inference-action}.

\paragraph{Active programmatic inference.}
LLM-driven discovery and active-experimentation benchmarks use evaluators or chosen interventions to search program space or identify hidden systems~\citep{romera2024mathematical, yamada2025ai, yin2025investigating, wei2025codearc, geng2025large, choudhury2025bed, zhang2024probing}. Program-synthesis work similarly treats latent mechanisms as code, including learned world models~\citep{tang2024worldcoder}, scientific programs~\citep{shojaee2024llm}, and programmatic models of other agents~\citep{jha2025modeling}. \benchname{} combines these ideas under constrained intervention: the learner cannot query arbitrary target states, but can write probe opponents that induce informative trajectories and refine an executable hypothesis against behavioral mismatches. Concurrent work on executable reconstruction, such as \citep{yang2026programbench}, asks agents to rebuild full software projects from a compiled program and documentation using behavioral equivalence tests; \benchname{} instead studies policy-level reverse engineering, where the target is an interacting agent and informative inputs must be induced through opponent policies rather than queried directly.

\section{Limitations}
\label{subsec:limitations}

Our setting is a controlled proxy for behavioral inference. The targets are synthetic, fixed game policies rather than adaptive or stochastic agents. They do not adapt across interactions, conceal strategy, or react to probing. Real-world agents may respond to observation or intervention, making reverse engineering substantially harder.

Action distance is only a proxy for policy equivalence. It measures behavioral fidelity on visited states, but rare or adversarially induced states may matter disproportionately, and small action differences can still cause large downstream effects. More fundamentally, behavioral recovery is not always identifiable: distinct programs may induce near-identical behavior under finite interaction budgets. Exact source recovery can therefore be underdetermined, with the recovered policy representing one member of a behaviorally consistent equivalence class rather than a unique solution.

Finally, benchmark reliability is arena-dependent. Our robustness analysis shows that BattleSnake and Poker yield relatively stable rankings across runs and evaluators, while Halite, RoboCode, and RobotRumble exhibit substantially higher variance, requiring multi-run evaluation for fine-grained comparison. In addition, probing entangles reasoning, coding, debugging, and experimental design, making failure modes difficult to attribute to any single capability.

\section{Conclusion}
\label{subsec:conclusion}

\benchname{} studies whether LLM coding agents can recover executable policies from behavioral evidence. Across frontier models, recovery is substantial but far from solved.

Our results suggest three main conclusions. First, passive behavioral evidence is often surprisingly informative: even without intervention, models recover a substantial fraction of target behavior, indicating that many programmatic strategies leave strong behavioral signatures in trajectory data. Second, active probing helps only when agents can use it effectively. Probing requires identifying uncertainty, designing informative interventions, implementing valid probes, and integrating the resulting evidence into policy revisions. Third, exact recovery is not necessary for downstream utility. Even imperfect reconstructions improve counter-strategy generation, especially for weaker challengers, suggesting that strategically useful opponent models need not faithfully reproduce the target source code.

These findings position behavioral recovery of executable policies as a useful testbed for opponent modelling, policy interpretability, and experimental reasoning in LLM agents. Future work should extend this setting to adaptive agents that learn or strategically conceal behavior, richer domains such as web and scientific agents, and inference methods that recover distributions or equivalence classes of plausible policies rather than a single executable hypothesis.

\newpage
\bibliographystyle{abbrvnat}
\bibliography{bib/neurips2026_conference}

\newpage
\appendix

\begin{ack}
BR, SD, JS, SH, and MB acknowledge support by the Tübingen AI Center. BR acknowledges funding from the European Union's Horizon Europe research and innovation program under the Marie Sk\l{}odowska-Curie grant agreement No.\ 101151549. SD, JS, and SH thank the International Max Planck Research School for Intelligent Systems (IMPRS-IS) for support. JS and SH acknowledge funding by the Federal Ministry of Research, Technology and Space (BMFTR), FKZ: 16IS24085B. MB acknowledges Coefficient Giving funded by the Good Ventures Foundation. MB acknowledges funding by the Federal Ministry of Research, Technology and Space (BMFTR), FKZ: 16IS24079A. MB is a member of the Machine Learning Cluster of Excellence, funded by the Deutsche Forschungsgemeinschaft (DFG, German Research Foundation) under Germany’s Excellence Strategy – EXC number 2064/1 – Project number 390727645.
\end{ack}

\section*{Broader impact}
\label{subsec:broader-impact}
The techniques that yield interpretable opponent models for coordination and scientific-inference studies could also be used to reverse-engineer proprietary decision systems, or, if extended beyond programmatic targets, to infer hidden processes from human behavior. We limit the benchmark to synthetic game policies and release artifacts under terms that discourage adversarial deployment; broader applications would require additional safeguards.

\section*{Appendix Outline}
\begin{enumerate}[label=\Alph*.]
  \item \nameref{app:extended-related-work}
  \item \nameref{app:strategy-pool}
  \item \nameref{app:algorithm}
  \item \nameref{app:baseline}
  \item \nameref{app:distances}
  \item \nameref{app:extended-results}
  \item \nameref{sec:forward-pvp}
  \item \nameref{app:history-management}
  \item \nameref{app:noise-decomposition}
  \item \nameref{app:cost-analysis}
\end{enumerate}
\section{Extended Related Work}
\label{app:extended-related-work}

This appendix expands the related work discussion from Section~\ref{sec:related_works}. We group prior work into theory of mind in LLMs, opponent modelling, strategic-agent evaluation, LLM-generated policies, active experimentation, and program synthesis from behavior.

\paragraph{Theory of mind in LLMs.}
A large body of work asks whether models can infer the mental states of other agents in order to predict their behavior. Machine theory-of-mind (ToM) models introduced this question in a trajectory-based setting, learning to infer an agent's goals and future actions from observed behavior~\citep{rabinowitz2018machine}. More recent LLM work has mostly evaluated ToM through static or semi-static text tasks, including causal-template vignettes~\citep{gandhi2023understanding}, cooperative games with explicit belief states~\citep{li2023theory}, and inference-time methods that combine LLM hypothesis generation with Bayesian or model-based reasoning over latent mental states~\citep{kim2025hypothesis, zhang2025autotom, cross2025hypothetical}. A key critique is that many such benchmarks measure whether models can answer ToM-style questions, rather than whether inferred beliefs are used to improve decisions in an interaction~\citep{riemer2025position}. \benchname{} addresses this gap by evaluating whether recovered behavioral models transfer to action: Section~\ref{sec:inference-action} tests whether code inferred from behavior helps a challenger win against the target.

\paragraph{Opponent modelling and opponent shaping.}
Opponent modelling studies how an agent can infer another agent's policy from observed actions and use that model for decision-making~\citep{nashed2022survey}. Classical approaches include learning opponent-aware update rules in small games~\citep{foerster2017learning, lu2022model}, scaling opponent shaping to higher-dimensional settings with temporally extended actions~\citep{khan2023scaling}, and unifying shaping methods under advantage-alignment principles~\citep{duque2024advantage}. Closer to our setting, \citet{jing2024opponent} combine in-context learning with decision-time search over possible opponent policies in poker. Recent work has extended opponent modelling to LLM agents, including explicit opponent models in social-deduction games~\citep{yu2025llm} and inference-time opponent simulation for repeated negotiation~\citep{liu2026scaling}. These methods typically produce latent, textual, or otherwise opaque opponent representations. In contrast, \benchname{} requires the inferred opponent model to be an executable policy, making the hypothesis directly testable and inspectable.

\paragraph{LLMs in strategic and game-theoretic environments.}
A growing literature evaluates whether LLMs reason strategically when placed in multi-agent or game-theoretic interactions. Empirical studies of repeated games characterise behavioral signatures such as cooperation, retaliation, and forgiveness in frontier models~\citep{akata2025playing}. Broader benchmarks evaluate strategic reasoning across complete- and incomplete-information games~\citep{duan2024gtbench, costarelli2024gamebench}, while multi-agent society simulations study emergent group-level dynamics such as cooperation or collapse~\citep{piatti2024cooperate}. Recent infrastructure for scaling agent environments and evaluations~\citep{froger2025scaling} and multi-agent negotiation systems for engineering tasks~\citep{cheng2026quare} share the goal of evaluating agents in richer interactive settings. \benchname{} is complementary: rather than measuring how well an LLM plays a game directly, it measures whether the LLM can infer the programmatic policy of another player from behavioral evidence.

\paragraph{LLM-generated policies and code-space games.}
Our target policies are drawn from code-based game environments, where strategies are represented as executable programs. Recent work has shown that LLMs can generate and refine game-playing policies in code space, including self-play approaches that iteratively add stronger generated bots to a policy population~\citep{bachrach2025combining} and code-space response oracles that produce interpretable multi-agent policies~\citep{hennes2026codespaceresponseoraclesgenerating}. These methods study the forward direction: generating policies that play well. \benchname{} studies the inverse direction: recovering a hidden policy's decision program from behavioral traces and controlled interactions.

\paragraph{Active experimentation and scientific discovery.}
\benchname{} is also related to LLM-driven scientific discovery, where models propose hypotheses, run experiments, and use automated evaluators to search for solutions. Examples include program-search systems for mathematical discovery~\citep{romera2024mathematical} and end-to-end agentic pipelines that generate hypotheses, execute experiments, and write up findings~\citep{yamada2025ai}. Active experimentation benchmarks more directly test whether LLMs can identify hidden systems by choosing informative interventions or queries~\citep{yin2025investigating, wei2025codearc, geng2025large}. Related work applies Bayesian experimental design to LLM-based question selection~\citep{choudhury2025bed} or uses entity-deduction games to measure multi-turn information gathering~\citep{zhang2024probing}. These works motivate our active-probing protocol: interventions can reveal distinctions that passive observations may not expose. The difference is that our interventions are constrained by the environment. The learner cannot query arbitrary target states, but must write opponent policies whose gameplay induces informative trajectories.

\paragraph{Program synthesis from behavior.}
Finally, \benchname{} connects to program-synthesis approaches that represent latent mechanisms as code. WorldCoder builds world models by writing code and interacting with an environment~\citep{tang2024worldcoder}. LLM-SR frames scientific equation discovery as programming with LLMs~\citep{shojaee2024llm}. Most directly related, \citet{jha2025modeling} model others' minds as code by generating candidate programs and scoring them against behavioral evidence. \benchname{} adopts the same broad premise that code is a useful representation for latent decision procedures, but changes the evaluation regime. Rather than selecting among program hypotheses from a passive batch, the learner iteratively edits a single executable policy, can actively probe the target through custom opponents, and is evaluated on held-out behavioral states as well as downstream strategic utility.
\section{Strategy Pool Construction}
\label{app:strategy-pool}

Our inverse strategy benchmark is built from a large corpus of strategies collected from public CodeClash tournament artifacts ~\cite{yang2025codeclash}. The goal of the construction pipeline is to obtain, for each game, a pool of executable and behaviorally diverse strategies that span a range of coding styles, structural complexity, and competitive strength. This appendix summarizes the data source, filtering pipeline, and pool selection procedure.

\begin{table}[h]
\centering
\small
\caption{Number of strategies retained at each stage of the strategy construction pipeline. Structural filtering removes invalid submissions, deduplication retains one strategy per similarity cluster, stratified selection builds the candidate pool, and Elo calibration defines the final target set.}
\label{tab:corpus-overview}
\setlength{\tabcolsep}{4pt}
\renewcommand{\arraystretch}{1.1}

\resizebox{\linewidth}{!}{%
\begin{tabular}{l 
l
>{\columncolor{blue!5}}r
>{\columncolor{blue!8}}r
>{\columncolor{blue!11}}r
>{\columncolor{blue!14}}r
}
\toprule
Game & Language & Raw corpus & Structural filtering & Deduplication + selection & Elo calibrated targets \\
\midrule
BattleSnake & Python & 4,372 & 438 & 40 & 15  \\
Halite & C & 1,743 & 622 & 40 & 15  \\
Poker & Python & 3,064 & 477 & 40 & 15  \\
RoboCode & Java & 1,676 & 522 & 40 & 15 \\
RobotRumble & JavaScript & 3,603 & 746 & 40 & 15 \\
\midrule
Total & --  & 14,458 & 2,805 & 200 & 75 \\
\bottomrule
\end{tabular}%
}
\end{table}

\subsection{Source Corpus}
\label{app:source-corpus}

We crawl strategy code from tournament artifacts hosted on the CodeClash website.\footnote{\url{https://codeclash.ai/}} We extract the final submitted strategy files for each participating model and game. The resulting corpus contains 14{,}891 raw strategy files across 1{,}901 tournament artifacts, covering six games and eight source models. Table~\ref{tab:corpus-overview} summarizes the corpus by game.

The six games span multiple programming languages and game mechanics: BattleSnake and Poker in Python, CoreWar in Redcode, Halite in C, RoboCode in Java, and RobotRumble in JavaScript.

\subsection{Filtering and Selection Pipeline}
\label{app:filtering-pipeline}

We construct the final strategy pools using a three stage pipeline: structural filtering, deduplication and stratified selection, and Elo calibration. The pipeline is game agnostic at the top level, with game specific validators and feature extractors used where needed.

\paragraph{Stage 1: Structural filtering.}
We first remove strategies that are clearly unusable without running full game simulations. A strategy must satisfy basic structural constraints such as the presence of the expected submission file, minimum code length, valid syntax, required handlers, and a valid runtime entrypoint. Additional game specific checks are applied when necessary. This stage reduces the raw corpus from 14{,}891 files to 3{,}314 structurally valid strategies.

\paragraph{Stage 2: Deduplication and candidate selection.}
Many CodeClash submissions are near duplicates, for example minor edits of the same underlying strategy or repeated generations with small formatting changes. To avoid over representing such families, we cluster strategies using cosine similarity over language appropriate feature representations. For Python games we use AST based feature vectors that capture node distributions, call patterns, control flow, and selected algorithmic motifs. For the other languages we use a combination of keyword indicators and hashed line shingles to represent structural similarity.

Within each game, we greedily construct a set of representatives by sorting strategies by a complexity score and keeping a new representative only when its similarity to all existing representatives is below a game specific threshold. The threshold is calibrated automatically from the empirical similarity distribution for that game. This yields 1{,}242 representative clusters across the six games.

Each representative is assigned a composite complexity score intended to capture structural richness rather than empirical win rate. The score combines factors such as code length, number of functions, nesting depth, branching structure, use of algorithmic patterns, and data structure diversity, with game specific feature extraction where needed. We then bucket strategies into low, medium, and high complexity tiers by percentile within each game.

To form the candidate pool for each game, we perform deterministic stratified sampling across source model and complexity tier. This encourages diversity both in where the strategy came from and in how sophisticated its code structure appears. For each game, we select 40 candidates. We exclude CoreWar from the final benchmark because its action representation does not admit the action-distance comparison our evaluation requires.

\paragraph{Validation after selection.}
After candidate selection, we apply a second validation gate to the populated pool to make sure the strategies run in the environment. We do this via self play inside the game Docker environment. If a selected strategy is found to be broken, it is replaced by the next available spare from the same deduplicated candidate set. This post selection validation is important because some failures, such as missing runtime assumptions or compilation issues, are not always detected by initial structural checks alone.

\paragraph{Stage 3: Elo calibration and target selection.}
\label{app:elo-calibration}
The 40 selected candidates for each arena are rated using a Swiss-system Elo tournament. In each round, strategies are paired against opponents with similar current ratings, and Elo is updated from head-to-head outcomes. We use 15 rounds with 10 simulations per match.

After rating, the 15 highest-Elo strategies per arena are designated as hidden targets. We restrict targets to the strongest policies because lower-rated candidates frequently exhibit failure modes that make them uninteresting as inverse-learning targets: their decision rule may be trivial (a single conditional or a constant action) or pathologically degenerate (for example, always folding pre-flop in poker, which produces deterministic play that loses predictably without exposing strategic structure). Restricting to the top tier avoids both pitfalls and provides headroom as frontier models continue to improve.

The remaining 39 strategies in each arena serve as the opponent pool. For each target, 20 opponents are sampled per tournament round to generate the behavioral traces the learner observes.

\begin{table}[h]
\centering
% \small
\caption{Summary of Elo calibration tournaments for the 40 selected strategies in each game, ordered by Elo spread (highest to lowest).}
\label{tab:elo-stats}
\begin{tabular}{lrr}
\toprule
Game & Matches & Elo range \\
\midrule
\rowcolor{blue!15} Poker & 300 & 1285--1642 \\
\rowcolor{blue!12} RobotRumble & 300 & 1365--1692 \\
\rowcolor{blue!9} Halite & 300 & 1334--1652 \\
\rowcolor{blue!6} BattleSnake & 300 & 1298--1574 \\
\rowcolor{blue!3} RoboCode & 300 & 1412--1584 \\
\midrule
Total & 1,500 &  \\
\bottomrule
\end{tabular}
\end{table}

\section{Inference Strategy Loop: Full Pseudocode}
\label{app:algorithm}

Algorithm~\ref{alg:inverse-strategy-expanded} gives the full inverse-strategy procedure. It makes explicit three implementation details that are otherwise hidden inside the refinement loop.

\paragraph{Mismatch set $\mathcal{M}_r$.}
At the start of each round, we evaluate the previous hypothesis $\hat{\pi}_{r-1}$ against the current-round trajectories $\mathcal{T}_r$ and collect the (state, predicted action, target action) tuples on which the two disagree. The agent receives $\mathcal{M}_r$ alongside the raw traces as log files that can be read, giving it a focused signal of where its code is currently wrong without forcing it to re-derive the disagreements from logs.

\paragraph{Step budget.}
Refinement within a round runs until a per-round step budget is exhausted. A step is any tool call: editing a file, running an evaluation script, reading a log, or submitting a probe. This bounds the compute the agent can spend per round and forces it to allocate effort between exploration and refinement.

\paragraph{Probe-edit interleaving.}
The probe budget $B$ is separate from the step budget. Each probe costs one unit of $B$, triggers a new \textsc{Interact} call against $\pi^*$, appends the resulting trace to $\mathcal{T}_r$, and recomputes $\mathcal{M}_r$. Crucially, probes are issued from inside the same refinement loop as edits: the agent can write a probe, observe the target's response, and incorporate that response into its next edit within the same round.

\begin{algorithm}[t]
\caption{Inverse Strategy}
\label{alg:inverse-strategy-expanded}
\begin{algorithmic}[1]
\Require Black-box target $\pi^*$, opponent pool $\mathcal{O}$, rounds $R$, probe budget $B$
\Ensure Executable hypothesis $\hat{\pi}$
\State Initialize $\hat{\pi}_0$
\For{$r = 1, \ldots, R$}
    \State \textbf{Observe:} Sample opponents $O_r \subset \mathcal{O}$; collect traces $\mathcal{T}_r \gets \textsc{Interact}(\pi^*, O_r)$
    \State \textbf{Evaluate:} $\mathcal{M}_r \gets \{(s_t, \hat{\pi}_{r-1}(s_t), \pi^*(s_t)) : s_t \in \mathcal{T}_r,\; \hat{\pi}_{r-1}(s_t) \neq \pi^*(s_t)\}$
    \State \textbf{Refine:} $\hat{\pi}_r \gets \hat{\pi}_{r-1}$; $b_r \gets 0$
    \Repeat \Comment{agent editing steps}
        \State $\hat{\pi}_r \gets \textsc{Edit}(\hat{\pi}_r,\; \mathcal{T}_{1:r},\; \mathcal{M}_r)$
        \If{agent requests probe \textbf{and} $b_r < B$} \Comment{active probing}
            \State Write probe $o$; $\tau \gets \textsc{Interact}(\pi^*, o)$; $b_r \gets b_r{+}1$
            \State $\mathcal{T}_r \gets \mathcal{T}_r \cup \{\tau\}$; recompute $\mathcal{M}_r$
        \EndIf
    \Until{step budget exhausted}
\EndFor
\State \Return $\hat{\pi}_R$
\end{algorithmic}
\end{algorithm}
\section{Baselines}
\label{app:baseline}

We include three baselines to isolate the contributions of different components of the inverse-strategy pipeline: (1)~a pool-mean baseline that requires no learning, (2)~a description baseline that replaces raw traces with natural-language summaries~\citep{hennes2026codespaceresponseoraclesgenerating}, and (3)~a Bayesian Program Inference (BPI) baseline that replaces iterative refinement with best-of-$N$ hypothesis selection~\citep{jha2025modeling}.

\subsection{Pool-mean baseline}
\label{app:baseline-pool-mean}

The pool-mean baseline estimates the expected recovery achievable by randomly selecting an existing strategy from the pool without any learning. For each target, we sample 5 strategies uniformly at random from the 40-strategy pool (excluding the target) and evaluate each against the target by computing the action distance. The best (lowest) distance among the 5 samples is recorded. We then compute the normalised recovery score for model evaluation. The reported pool-mean score is the average over all targets. This baseline quantifies how much of the behavioral gap can be closed by chance similarity between pool strategies and the target, without any inference or code synthesis.

\subsection{Description-only baseliene}
\label{app:baseline-nl-obs}
Description tests whether raw behavioral traces can be replaced by a compact natural-language observation channel, inspired by description-input code generation~\citep{hennes2026codespaceresponseoraclesgenerating}. After each round, a GPT-5.4-mini summariser converts sampled target behavior into a structured description of apparent decision rules. The learner receives only these summaries and refines its hypothesis code over $R{=}5$ rounds. It has no access to raw trace files and no probing budget. This compares raw-trace refinement against a passive summarise-then-refine pipeline.

\paragraph{Summariser.}
For each round, a GPT-5.4-mini summariser receives a sampled set of mismatch examples, i.e. state--action cases where the learner's action differs from the target's action. It produces a concise structured report of the target's apparent decision rules, including state-dependent behaviors, opponent-specific responses, threshold-like patterns, and consistent action preferences. The report is written as a text file in the learner's observation directory and remains available in later rounds; the underlying behavioral traces are not exposed.

\paragraph{Learner configuration.}
The learner uses the same initial code template, system prompt, editing interface, and round structure as the main method, but active probing is disabled. It runs for $R{=}5$ rounds with 20 passive opponents per round, matching the main evaluation protocol.

\subsection{Bayesian Program Inference baseline}
\label{app:baseline-bpi}

Bayesian Program Inference (BPI) tests whether iterative refinement can be replaced by passive best-of-$N$ program inference, inspired by program-hypothesis scoring methods~\citep{jha2025modeling}. BPI uses raw behavioral traces, generates several independent executable hypotheses, and selects the candidate that best explains the observed target behavior under a distance-based posterior score.

\paragraph{Procedure.}
BPI runs one passive simulation round between the target and the fixed set of 20 opponents. It then spawns $N{=}5$ independent LLM editing sessions, each initialized with the same code template and given the game specification, policy API, and a bootstrap-resampled subset of the observed behavioral traces. Each session produces one candidate policy, with no shared memory or intermediate outputs across sessions. Candidates are validated using the same sandbox and API checks as the main learner. The highest-scoring candidate is selected as the final recovered policy.

\paragraph{Hypothesis diversity.}
To reduce collapse to near-duplicate samples, candidate generation uses two diversity mechanisms. First, each candidate receives a different bootstrap-resampled subset of the behavioral traces, so different opponent interactions can drive different hypotheses. Second, each candidate receives an approach-conditioning hint that encourages a distinct explanatory lens, such as minimal if--then rules, spatial or positional patterns, state-dependent mode switching, opponent-adaptive behavior, or resource/value optimisation. These mechanisms affect only generation; all candidates are scored on the same full observed trace set.

\paragraph{Distance-based posterior score.}
For candidate $i$, let $D_i = |\mathcal{O}|^{-1}\sum_{o \in \mathcal{O}} d(i,o)$ be its mean action distance to the target across the observed opponent set $\mathcal{O}$. We score candidates as
\[
    s_i = -\beta D_i - \alpha c_i,
\]
where $c_i$ is the number of non-trivial lines in the candidate code divided by 100. We use $\beta{=}10$ and $\alpha{=}0.01$. This score corresponds to a distance-based likelihood with a weak complexity prior favouring shorter programs among candidates with similar behavioral fit. The submitted policy is the MAP candidate, i.e. $\arg\max_i s_i$.

\section{Per-Game Action Distances}
\label{app:distances}

This appendix gives the full definition of each per-game action distance $d_g : \mathcal{A}_g \times \mathcal{A}_g \to [0, 1]$ summarized in Table~\ref{tab:per-game-distance}. All five distances satisfy the conventions established in Sec.~\ref{setting}: $d_g(a, a) = 0$, $d_g$ is symmetric, $d_g$ equals $1$ when either argument is missing or invalid, and aggregation over action components is normalized so that scores remain comparable across states of differing dimensionality.

The recurring design choice across arenas is whether a mismatch is binary or graded. Where the action space is genuinely discrete and unordered (e.g., a single direction in BattleSnake), a binary distance is the only honest summary: "almost right" has no defensible metric meaning. Where the action carries quantitative content (raise sizes in poker, turn rates in RoboCode, per-cell move counts in Halite), a continuous distance is essential. The strategies we study encode behaviors in those quantities, and ``right behavior, wrong magnitude'' is a qualitatively different failure mode from ``completely wrong.'' Collapsing the two onto a binary scale would flatten partial recovery into a single bit and make the metric blind to most of the recovery signal we care about.

Importantly, we always use one policy (the target) to roll out the actions against an opponent. For each state, we compare the action predicted by the agent's policy and compute the distance between the two. This ensures that we always compare actions that are directly comparable because they were taken in the exact same situation. Rolling out games using the target policy also guarantees that the learned policy is functionally similar to the target if it shows a small average distance, because it prevents cheating by only visiting states where the correct action is easy to infer. Conceptually, this is similar to the differences between forward and backward Kullback-Leibler Divergence.

\subsection{BattleSnake~\citep{chung2020battlesnake}}
\label{app:dist:battlesnake}

In BattleSnake, the action space is $\mathcal{A}_{\text{BS}} = \{\textsc{up}, \textsc{down}, \textsc{left}, \textsc{right}\}$: at each turn the snake selects exactly one of four cardinal directions. There is no natural ordering on this set: opposite directions are typically lethal (the snake collides with itself), but adjacent directions can have radically different outcomes depending on board state. We therefore use a binary distance,
\begin{equation}
    \label{eq:dist-battlesnake}
    d_{\text{BS}}(a_1, a_2) \;=\; \mathbf{1}[a_1 \neq a_2].
\end{equation}
A continuous relaxation (e.g., $0.5$ for adjacent directions, $1$ for opposite) was considered and rejected: the same direction can be optimal or fatal depending on the local configuration, so any geometric weighting would impose structure that the game does not have.

\subsection{Halite~\citep{truell2016halite}}
\label{app:dist:halite}

In Halite, an action $a \in \mathcal{A}_{\text{Hal}}$ is a partial assignment from owned cells to moves: each ship or factory under the player's control receives one of a small set of commands (e.g., \textsc{stay}, $\{\textsc{n}, \textsc{s}, \textsc{e}, \textsc{w}\}$, \textsc{spawn}, \textsc{construct}). Because the number of owned cells varies across game states, we cannot simply compare actions component-wise. Instead, we score each action over the union of cells assigned by either policy.

Let $\mathcal{C}(a) \subseteq \mathcal{C}_t$ denote the set of cells to which $a$ assigns a command in state $s_t$, and let $\mathcal{C}_{\cup} = \mathcal{C}(a_1) \cup \mathcal{C}(a_2)$. For any cell $c \notin \mathcal{C}(a)$ we set $a[c] = \varnothing$ (no command), treated as a distinct value from any explicit command. The distance is the fraction of cells in the union on which the two policies disagree:
\begin{equation}
    \label{eq:dist-halite}
    d_{\text{Hal}}(a_1, a_2) \;=\; \frac{1}{|\mathcal{C}_{\cup}|} \sum_{c \in \mathcal{C}_{\cup}} \mathbf{1}[a_1[c] \neq a_2[c]],
\end{equation}
with $d_{\text{Hal}} = 0$ when both policies issue empty commands ($\mathcal{C}_{\cup} = \emptyset$). Averaging over $|\mathcal{C}_{\cup}|$ (rather than over $|\mathcal{C}_t|$) penalizes disagreements on commanded cells without rewarding either policy for declining to issue commands. Cells commanded by only one policy are scored as full mismatches, on the same principle: an omission in one policy is a substantive deviation from the other, not a free pass.

\subsection{Poker (HuskyBench)~\citep{kumar2025huskybench}}
\label{app:dist:poker}

In Poker, an action is one of $\{\textsc{fold}, \textsc{check}, \textsc{call}, \textsc{raise}{:}r\}$ with $r \in (0, 1]$ encoding the raise size as a fraction of the actor's chip stack at the decision point. The metric design separates two effects: the categorical decision to fold versus continue, and (for non-folds) the continuous magnitude of commitment.

Folds are handled categorically. We set $d_{\text{HB}}(\textsc{fold}, \textsc{fold}) = 0$ and $d_{\text{HB}}(\textsc{fold}, a) = 1$ for any non-fold $a$. This reflects the qualitative gap between folding (conceding the hand) and any continuation: folds and non-folds have no reasonable interpolation. We also identify $\textsc{check}$ and $\textsc{call}$ for distance purposes, treating both as a zero-commitment $\textsc{raise}{:}0$. The two are mutually exclusive in any given state (the legal action depends on whether the actor faces a bet), so distinguishing them in the metric would only add noise.

For two non-fold actions, we compute a stack-aware log-commitment distance. Let $S$ be the actor's chip stack at the decision point and let $\rho(a) \in [0, 1]$ be the raise ratio ($0$ for $\textsc{check}$/$\textsc{call}$, $r$ for $\textsc{raise}{:}r$). Define the commitment function
\[
    c(\rho; S) \;=\; \frac{\log(1 + \rho S)}{\log(1 + S)}.
\]
The distance between non-fold actions is then
\begin{equation}
    \label{eq:dist-poker}
    d_{\text{HB}}(a_1, a_2) \;=\; \bigl| c(\rho(a_1); S) - c(\rho(a_2); S) \bigr|.
\end{equation}
The log scale reflects two empirical facts about poker. First, the strategic distance between a 1\%-of-stack raise and a 5\%-of-stack raise is larger than between a 50\%-of-stack raise and a 55\%-of-stack raise, even though the absolute differences are equal. Second, the smallest legal raise should be meaningfully separated from a passive $\textsc{check}$/$\textsc{call}$ rather than collapsing to nearly-zero commitment. Normalizing by $\log(1 + S)$ rather than a constant guarantees $d_{\text{HB}} \in [0, 1]$ in every state: an all-in is at distance $1$ from a $\textsc{check}$ regardless of stack depth.

\subsection{RoboCode~\citep{hartness2004robocode}}
\label{app:dist:robocode}

In RoboCode, an action is a five-dimensional continuous control vector $a \in \mathbb{R}^5$ with components $(\text{velocity}, \text{turn\_body}, \text{turn\_gun}, \text{turn\_radar}, \text{fire\_power})$. The components are physically heterogeneous: a difference of $1$ in $\text{fire\_power}$ is qualitatively different from a difference of $1$ in $\text{turn\_radar}$. We therefore normalize each component by its valid range $R^{(i)}$ before averaging:
\begin{equation}
    \label{eq:dist-robocode}
    d_{\text{RC}}(a_1, a_2) \;=\; \frac{1}{5} \sum_{i=1}^{5} \min\!\left( \frac{|a_1^{(i)} - a_2^{(i)}|}{R^{(i)}},\; 1 \right),
\end{equation}
with component ranges taken from the RoboCode specification:
\[
\begin{array}{lll}
    R^{\text{velocity}} = 16 \;\;([-8, 8]), &
    R^{\text{turn\_body}} = 20 \;\;([-10, 10]), &
    R^{\text{turn\_gun}} = 40 \;\;([-20, 20]), \\[2pt]
    R^{\text{turn\_radar}} = 90 \;\;([-45, 45]), &
    R^{\text{fire\_power}} = 3 \;\;([0, 3]). &
\end{array}
\]
The inner $\min$ clips per-component differences at $1$, so that an out-of-range output (which the game engine would otherwise clip silently) cannot push the overall distance above $1$. Equal weighting across components reflects that we have no principled basis for declaring one channel more strategically important than another; targets in our pool exercise all five.

\subsection{RobotRumble~\citep{outkine2020robotrumble}}
\label{app:dist:robotrumble}

In RobotRumble, a player controls multiple units simultaneously, and an action is a list of (unit\_id, $(\textsc{type}, \textsc{dir})$) pairs, where $\textsc{type} \in \{\textsc{Move}, \textsc{Attack}\}$ and $\textsc{dir}$ is one of the eight unit-step directions. Disagreement on $\textsc{type}$ is a categorically larger error than disagreement on $\textsc{dir}$ alone: moving and attacking change different properties of the world, while a wrong direction within the same action type is a localized tactical mistake. We encode this asymmetry in a three-level per-unit distance.

Let $U = \mathcal{U}(a_1) \cup \mathcal{U}(a_2)$ be the set of unit identifiers commanded by either action. For a unit $u \in U$, define
\[
d_u(a_1, a_2) \;=\; \begin{cases}
0   & \text{if } a_1[u] = a_2[u],\\
0.5 & \text{if } \textsc{type}(a_1[u]) = \textsc{type}(a_2[u]) \text{ and both actions are present},\\
1   & \text{otherwise (different type, or one action missing for } u\text{)},
\end{cases}
\]
The full distance averages over the set $U$ of units commanded in the state:
\begin{equation}
    \label{eq:dist-robotrumble}
    d_{\text{RR}}(a_1, a_2) \;=\; \frac{1}{|U|} \sum_{u \in U} d_u(a_1, a_2).
\end{equation}
where $a[u]$ is taken to be missing if $u \notin \mathcal{U}(a)$. The action distance is the average over $U$:
\begin{equation}
    \label{eq:dist-robotrumble}
    d_{\text{RR}}(a_1, a_2) \;=\; \frac{1}{|U|} \sum_{u \in U} d_u(a_1, a_2),
\end{equation}
with $d_{\text{RR}} = 0$ when both actions are empty ($U = \emptyset$). Failing to issue a command for a unit that the other policy commands is treated as a maximally different choice: in RobotRumble, idleness is itself a strategic decision and is rarely correct.

\section{Extended Results}
\label{app:extended-results}

\subsection{Per-game strategy recovery}
\label{app:per-game-recovery}

\begin{figure}[t]
    \centering
    \includegraphics[width=0.9\linewidth]{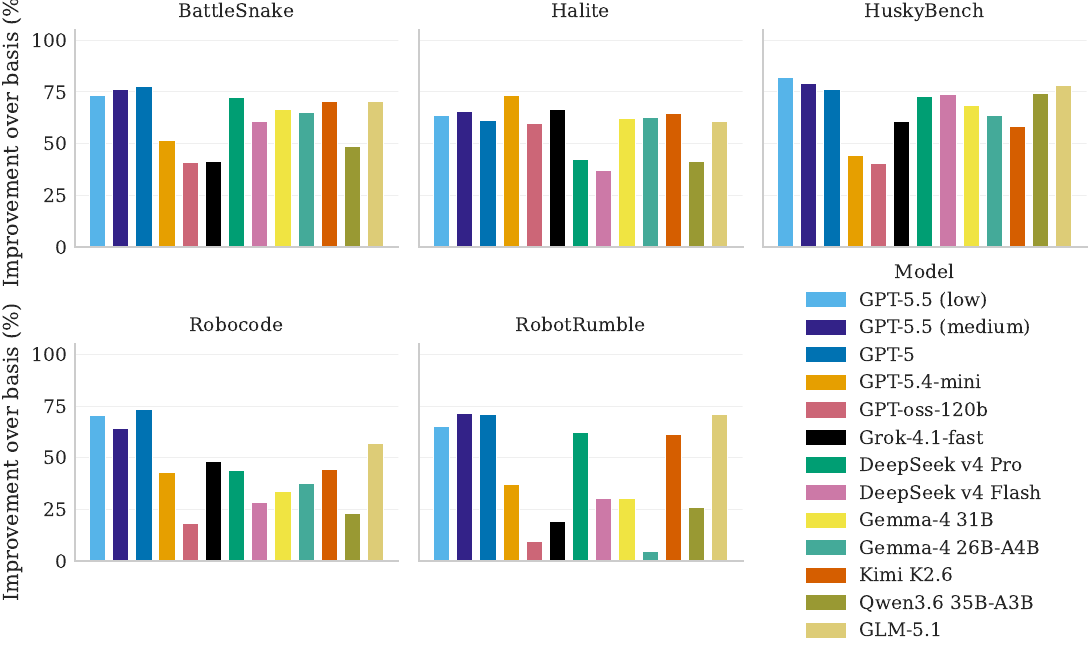}
    \caption{Per-game normalised recovery score for all models. Performance rankings vary across environments.}
    \label{fig:per-game-recovery}
\end{figure}
Figure~\ref{fig:per-game-recovery} breaks down the aggregate recovery scores from Section~\ref{subsec:strategy-recovery} by game environment. Model rankings are not uniform across arenas, revealing environment-specific strengths. GPT-5.5-Codex-high achieves the highest overall recovery (83\%) and leads or co-leads in every arena.

\paragraph{BattleSnake.} GPT-5.5-Codex-high leads (81\%), followed by GPT-5 and GPT-5.5-Codex-low (78\%), GPT-5.5-medium (76\%), and GPT-5.5-low (74\%). Kimi~K2.6 (71\%) and DeepSeek~v4~Pro (72\%) form a strong second tier.

\paragraph{Halite.} GPT-5.5-Codex-high dominates (82\%), followed by GPT-5.4-mini (73\%), GPT-5.5-Codex-low (68\%), Gemma-4~26B-A4B and Grok-4.1-fast (67\%). GPT-5 (61\%) and DeepSeek~v4~Flash (40\%) trail.

\paragraph{Poker.} This is the easiest arena overall. GPT-5.5-Codex-high leads (87\%), followed by GPT-5.5-Codex-low (84\%), GPT-5.5-low (82\%), and GPT-5.5-medium (79\%). DeepSeek~v4~Flash and Qwen3.6~35B reach 74\%.

\paragraph{RoboCode.} GPT-5.5-Codex-high leads (74\%), followed by GPT-5 (73\%) and GPT-5.5-low (71\%). GPT-5.5-medium and GPT-5.5-Codex-low reach 62--64\%. GPT-oss-120b recovers only 18\%.

\paragraph{RobotRumble.} This arena shows the widest model spread. GPT-5.5-Codex-high dominates (89\%), followed by GPT-5.5-Codex-low (82\%). GPT-5, GPT-5.5-medium, and GLM-5.1 cluster around 71\%. Gemma-4~26B-A4B trails at 5\%.

\subsection{Probing statistics}
\label{app:probing-stats}

\begin{figure}[h]
    \centering
    \includegraphics[width=0.95\linewidth]{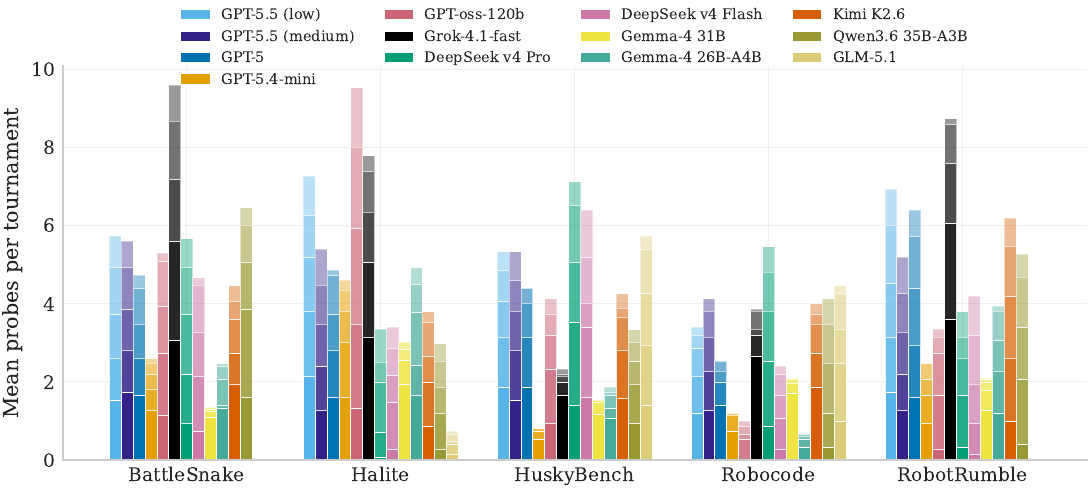}
    \caption{Mean number of probes executed per tournament, grouped by game. Each bar is segmented by round (darker = earlier rounds), showing the per-round breakdown within each model$\times$game combination.}
    \label{fig:probes-per-game}
\end{figure}

\begin{figure}[h]
    \centering
\includegraphics[width=0.95\linewidth]{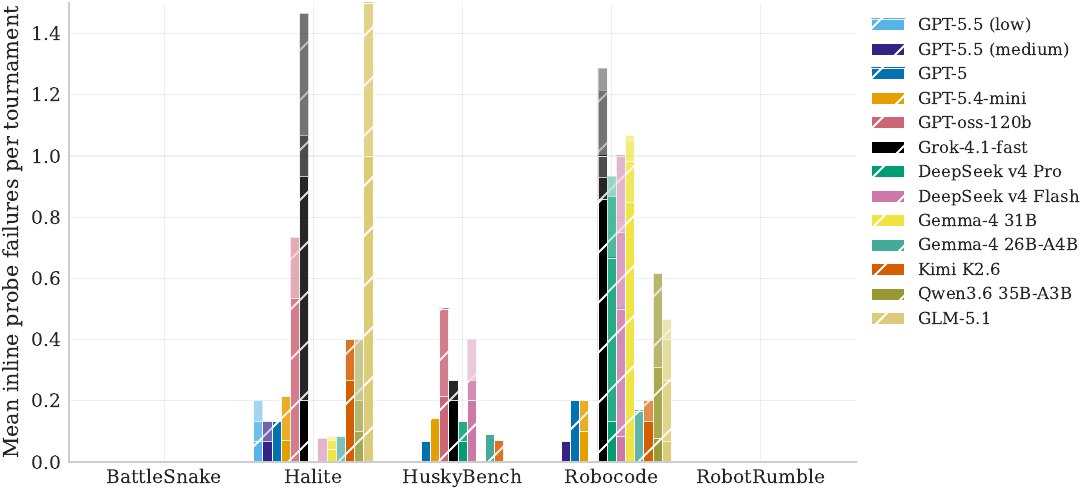}
    \caption{Mean inline probe failures per tournament. Each bar is segmented by round (darker = earlier rounds), showing when failures occur during refinement. Hatched bars indicate failed probe attempts (invalid code, sandbox errors, or timeouts). Failures concentrate in Halite, RoboCode, and RobotRumble; Grok-4.1-fast and GLM-5.1 show the highest rates in Halite, while DeepSeek~v4~Pro and Grok-4.1-fast dominate RoboCode failures. GPT-5.5 variants (including Codex) achieve near-zero failure rates across all arenas.}
    \label{fig:probe-failures}
\end{figure}

Figures~\ref{fig:probes-per-game} and~\ref{fig:probe-failures} characterise how models use their probing budget across games.

\paragraph{Probe usage.}
Figure~\ref{fig:probes-per-game} shows the mean number of probes executed per tournament, broken down by game and model. Each model has a budget of 5 probes per round. Probe usage is highest in early rounds and decreases in later rounds, as models converge on a hypothesis and shift effort toward code refinement rather than further experimentation. Some games (BattleSnake, Halite) tend to elicit more probing.

\paragraph{Probe failures.}
Figure~\ref{fig:probe-failures} shows the mean number of inline probe failures (invalid submissions, sandbox errors, or timeouts) per tournament. Weaker models exhibit higher failure rates. This confirms that probing is a tool-use capability: models must not only decide \emph{what} to probe but also produce valid, executable probe policies. Failed probes waste budget without yielding information, explaining why some models perform worse with probing enabled than without (Section~\ref{subsec:observation-intervention}).

\subsection{Probe effectiveness}
\label{app:probe-effectiveness}

To quantify whether active probing improves strategy recovery, we compare best action distance $D_{min}$ between Active (with probes) and Passive (no probes) conditions across the same four models.
We define the \emph{probe boost} as the reduction in distance afforded by probing: $\Delta = D_{min,~\text{passive}} - D_{min,~\text{active}}$, where positive values indicate that probing helped.

\begin{figure}[h]
    \centering
    \includegraphics[width=0.85\linewidth]{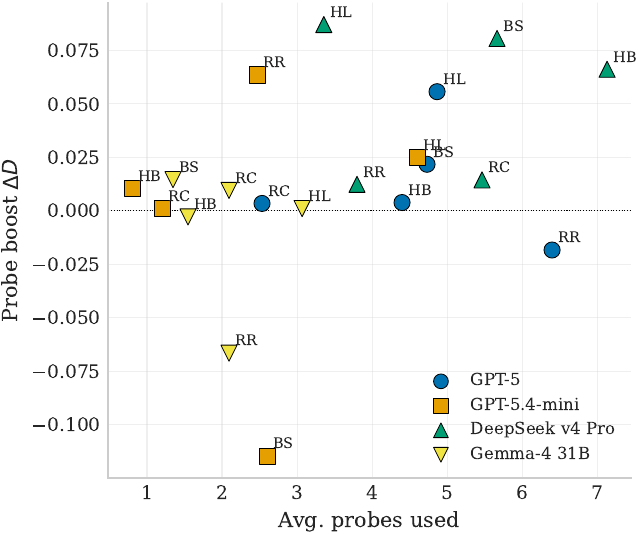}
    \caption{Probe boost vs.\ average probes used per tournament.
    Each point is one (model~$\times$~game) pair.
    Points above the dashed line indicate probing reduced action distance (helped recovery); 16 of 20 points are positive.
    Marker shape encodes model; game abbreviations are annotated (BS=BattleSnake, HL=Halite, HB=Poker, RC=Robocode, RR=RobotRumble).}
    \label{fig:probe-boost}
\end{figure}

Figure~\ref{fig:probe-boost} plots probe boost against the average number of probes consumed per tournament for each (model~$\times$~game) pair.
Across all 20 combinations, 16 out of 20 (80\%) show a positive boost, with an overall mean improvement of $+0.013$ ($\sigma{=}0.046$).
The effect is most pronounced for DeepSeek~v4~Pro, which benefits from probing in all five games (mean boost $+0.052$), with the largest gains in Halite ($+0.087$) and Poker ($+0.066$).
GPT-5 benefits in four of five games (mean $+0.013$), particularly in Halite ($+0.056$).
GPT-5.4-mini and Gemma-4~31B show near-zero mean benefit, with individual games exhibiting both positive and negative effects. This indicates that for weaker models, probing can occasionally hurt when it consumes budget without yielding actionable information.

\section{From Inference to Action}
\label{sec:forward-pvp}

The inverse-strategy pipeline recovers a target's decision procedure from behavioral traces and interaction with the target.
A natural question is whether knowing the target's strategy is useful: if an agent has access to a model of its opponent's logic, can it leverage that model to win more often?
This connects our pipeline to the broader question of theory of mind in competitive settings: whether understanding how an opponent makes decisions translates into a concrete strategic advantage.

We test this with a \emph{PvP} (player-versus-player) experiment.
A \emph{challenger} agent receives the game rules and must write a counter-strategy over multiple rounds of iterative editing, receiving game outcomes after each round.
The key manipulation is what the challenger knows about its opponent: nothing (blind), the code recovered by our inverse pipeline (recovered), or the opponent's actual source code (oracle).
If opponent intelligence helps, the recovered and oracle conditions should outperform blind; if the recovery is faithful, the recovered condition should approach the oracle.

\subsection{Protocol}
\label{sec:forward-pvp-protocol}

Each experiment instance pits two players against each other over $R{=}5$ rounds of $S{=}30$ simulations each.

\paragraph{Player~1 (target).}
Plays its original strategy code, held fixed throughout the tournament.
The target is not an LLM agent; it simply executes the source code that was the subject of the inverse-strategy evaluation.

\paragraph{Player~2 (challenger).}
An LLM-powered coding agent (Section~\ref{sec:agent}) that iteratively edits a counter-strategy.
After each round, the challenger observes the game results and may revise its code.
The challenger sees the standard game description prompt for the relevant game; the only variable across conditions is the presence and source of opponent intelligence.

We compare three levels of opponent information:

\begin{enumerate}[label=(\alph*), nosep, leftmargin=*]
    \item \textbf{Blind.} The challenger receives only the game rules. No opponent code is provided. This is the control baseline: any wins come from the challenger's intrinsic coding ability and its iterative adaptation to game outcomes.
    \item \textbf{Recovered.} The challenger receives the code recovered by our inverse-strategy pipeline (the best-round submission from the main inverse strategy evaluation). The code is injected into the game prompt as an ``opponent intelligence'' section.
    \item \textbf{Oracle.} The challenger receives the target's actual source code. This serves as a ceiling: if even perfect knowledge does not help, opponent modelling is uninformative for the game in question.
\end{enumerate}

\noindent In the recovered and oracle conditions, the opponent code is appended to the game description with the following prompt, deliberately worded to discourage brittle hard-coded exploits that would succeed against the exact target code but fail against minor perturbations:

\begin{tcolorbox}[
    enhanced,
    colback=promptpurple,
    colframe=promptborder,
    boxrule=0.6pt,
    arc=3pt,
    left=8pt, right=8pt, top=6pt, bottom=6pt,
    fontupper=\small\itshape,
    title={\small\bfseries\upshape Opponent Intelligence Prompt},
    coltitle=black,
    colbacktitle=promptpurple!80!promptborder,
]
You have obtained your opponent's strategy code: \texttt{[code]}.
Use this intelligence to inform your strategy.
A well-designed bot that accounts for the opponent's tendencies will outperform a fragile exploit.
Focus on writing robust, well-tested code first, then incorporate any insights from the opponent's logic as refinements.
\end{tcolorbox}

\subsection{Games, Targets, and Models}
\label{sec:forward-pvp-design}

We evaluate on four games from our benchmark: \textsc{BattleSnake}, \textsc{Halite}, \textsc{RoboCode} (Java oracle / Python recovered), and \textsc{RobotRumble}.
For each game, we select the top-5 targets by behavioral distance from the main inverse strategy (Section~\ref{sec:results-main}), i.e., the five targets whose strategies were most faithfully recovered.
Selecting well-recovered targets ensures that the recovered condition contains a meaningful approximation of the target's logic; poorly recovered targets would make the comparison uninformative.

Five challenger LLMs span a range of coding ability: GPT-5, GPT-5-mini, GPT-oss-120b, DeepSeek-v3.2, and Grok-4.1-fast.
Each model is used as the challenger in all three conditions against all 20 targets (5~per game~$\times$~4~games).
All models are evaluated with three independent tournament seeds. This yields $4 \times 3 \times 60 = 720$ tournament runs and approximately 81{,}000 simulations; all reported win rates are averaged across seeds.
Using models of varying strength allows us to test whether the value of opponent intelligence depends on the challenger's baseline capability.

\subsection{Metric}
\label{sec:forward-pvp-metric}

We measure the challenger's \emph{win rate} per round:
\begin{equation}
    \mathrm{WR}_r = \frac{s_{\text{challenger},r}}{s_{\text{challenger},r} + s_{\text{target},r}} \times 100\%,
\end{equation}
where $s_{\cdot,r}$ is the cumulative simulation score in round~$r$.
The primary summary statistic is the mean win rate over rounds $1$--$5$.
We define the \emph{win rate gain} as $\mathrm{WR}_{\text{condition}} - \mathrm{WR}_{\text{blind}}$, which isolates the contribution of opponent information from the challenger's baseline performance.
Specifically, \emph{oracle lift} $= \mathrm{WR}_{\text{oracle}} - \mathrm{WR}_{\text{blind}}$ and \emph{recovered lift} $= \mathrm{WR}_{\text{recovered}} - \mathrm{WR}_{\text{blind}}$.

\subsection{Hypotheses}
\label{sec:forward-pvp-hypotheses}

We test three hypotheses:

\begin{description}[style=unboxed, leftmargin=0pt]
    \item[H1: Opponent information $>$ Blind.]
    Knowing the opponent's strategy, whether the exact source code (oracle) or the code recovered by our inverse pipeline (recovered), helps the challenger win.

    \item[H2: Advantage decays over rounds.]
    The initial boost from opponent knowledge shrinks as the blind challenger iteratively improves from game feedback alone.
    A strong decay would suggest that opponent intelligence provides a warm-start advantage rather than a persistent structural edge.

    \item[H3: Weaker models benefit more.]
    Models with lower blind baselines derive larger advantages from opponent information.
    If the blind baseline is already high, there is less room for improvement; if it is low, opponent intelligence fills the gap.
\end{description}

\subsection{H1: Opponent Information Improves Win Rate}
\label{sec:forward-pvp-h1}

\begin{figure}[h]
    \centering
    \includegraphics[width=\linewidth]{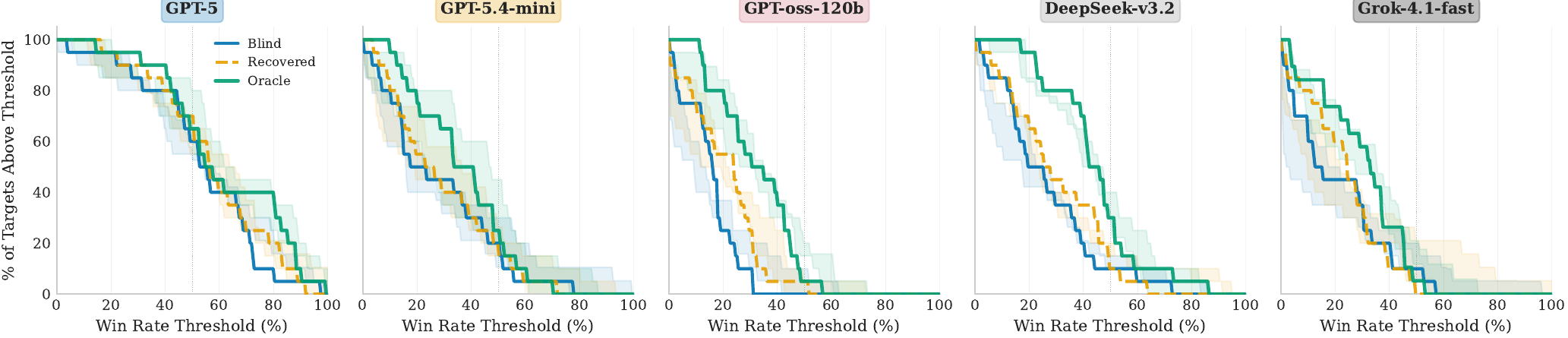}
    \caption{%
        Fraction of targets where the challenger achieves at least a given win rate, by condition.
        For GPT-5, the three curves nearly overlap, as blind performance is already strong.
        For weaker models, oracle and recovered curves shift visibly right of blind.
    }
    \label{fig:cdf}
\end{figure}

Figure~\ref{fig:pvp-main} (left) in the main paper shows that across all five models, oracle outperforms recovered, which in turn outperforms blind. The gap between recovered and oracle reveals that imperfect recovery captures only a fraction of the full intelligence value; nevertheless, the recovered condition consistently outperforms blind, establishing that the inverse-strategy pipeline produces actionable intelligence.
The survival-CDF view (Figure~\ref{fig:cdf}) reinforces this: for weaker models, oracle and recovered curves shift visibly right of blind, meaning these challengers beat more targets, and beat them more decisively, when armed with opponent information.

\subsection{H2: The Intel Advantage Decays for Strong Models but Persists for Weak Ones}
\label{sec:forward-pvp-h2}

\begin{figure}[h]
    \centering
    \includegraphics[width=0.65\linewidth]{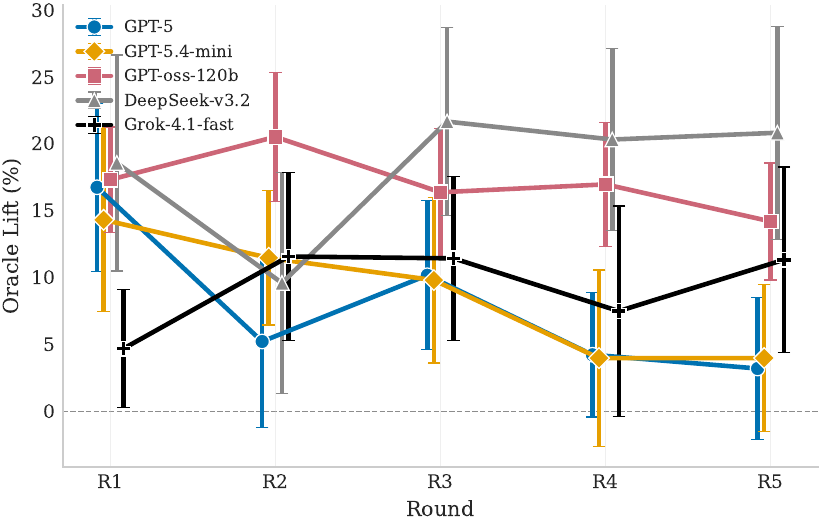}
    \caption{%
        Oracle lift (oracle minus blind win rate) across rounds, with $\pm$1 standard error bars.
        GPT-5 and GPT-5-mini start with large lifts that decay toward zero as their blind baselines catch up through iterative play.
        Weaker models show persistent lift because their blind performance barely improves.
    }
    \label{fig:r1-lift}
\end{figure}

Figure~\ref{fig:r1-lift} traces the oracle lift from Round~1 through Round~5.
Two regimes emerge.
GPT-5 and GPT-5.4-mini start with large lifts that shrink as their blind baselines catch up through iterative play. 
The picture reverses for weaker models: DeepSeek-v3.2, and Grok-4.1-fast show persistent oracle lift because their blind performance barely improves across rounds.

\subsection{H3: Weaker Models Benefit More from Intel}
\label{sec:forward-pvp-h3}

Figure~\ref{fig:pvp-main} (middle, right) from the main paper makes the relationship between baseline strength and win rate gain benefit explicit.
The negative slopes are significant ($p < 0.05$) for four of five models on both oracle lift and recovered lift (Table~\ref{tab:blind-vs-lift}), confirming that the lower the blind baseline, the larger the lift from opponent intelligence.

\begin{table}[h]
\centering
\small
\caption{Pearson correlation between blind win rate and win rate gain per model. Negative $r$ confirms that weaker blind baselines benefit more from opponent intelligence.}
\label{tab:blind-vs-lift}
\begin{tabular}{lcccc}
    \toprule
    & \multicolumn{2}{c}{Oracle lift} & \multicolumn{2}{c}{Recovered lift} \\
    \cmidrule(lr){2-3} \cmidrule(lr){4-5}
    Model & $r$ & $p$ & $r$ & $p$ \\
    \midrule
    GPT-5          & $-0.41$ & $0.069$        & $-0.53$ & $0.017$ \\
    GPT-5.4-mini   & $-0.74$ & ${<}0.001$   & $-0.55$ & $0.012$ \\
    GPT-oss-120b   & $-0.48$ & $0.033$   & $-0.43$ & $0.059$ \\
    DeepSeek-v3.2  & $-0.81$ & ${<}0.001$ & $-0.53$ & $0.016$ \\
    Grok-4.1-fast  & $-0.66$ & $0.002$ & $-0.66$ & $0.002$ \\
    \bottomrule
\end{tabular}
\end{table}

Crucially, the pattern holds for recovered lift (Table~\ref{tab:blind-vs-lift} right panel) as well as oracle lift, confirming that the recovered code, not just the ground-truth source, carries enough actionable signal to benefit weaker challengers.

\subsection{Per-Model Win-Rate Trajectories}
\label{sec:forward-pvp-per-model}

\begin{figure}[p]
    \centering
    \includegraphics[width=\linewidth]{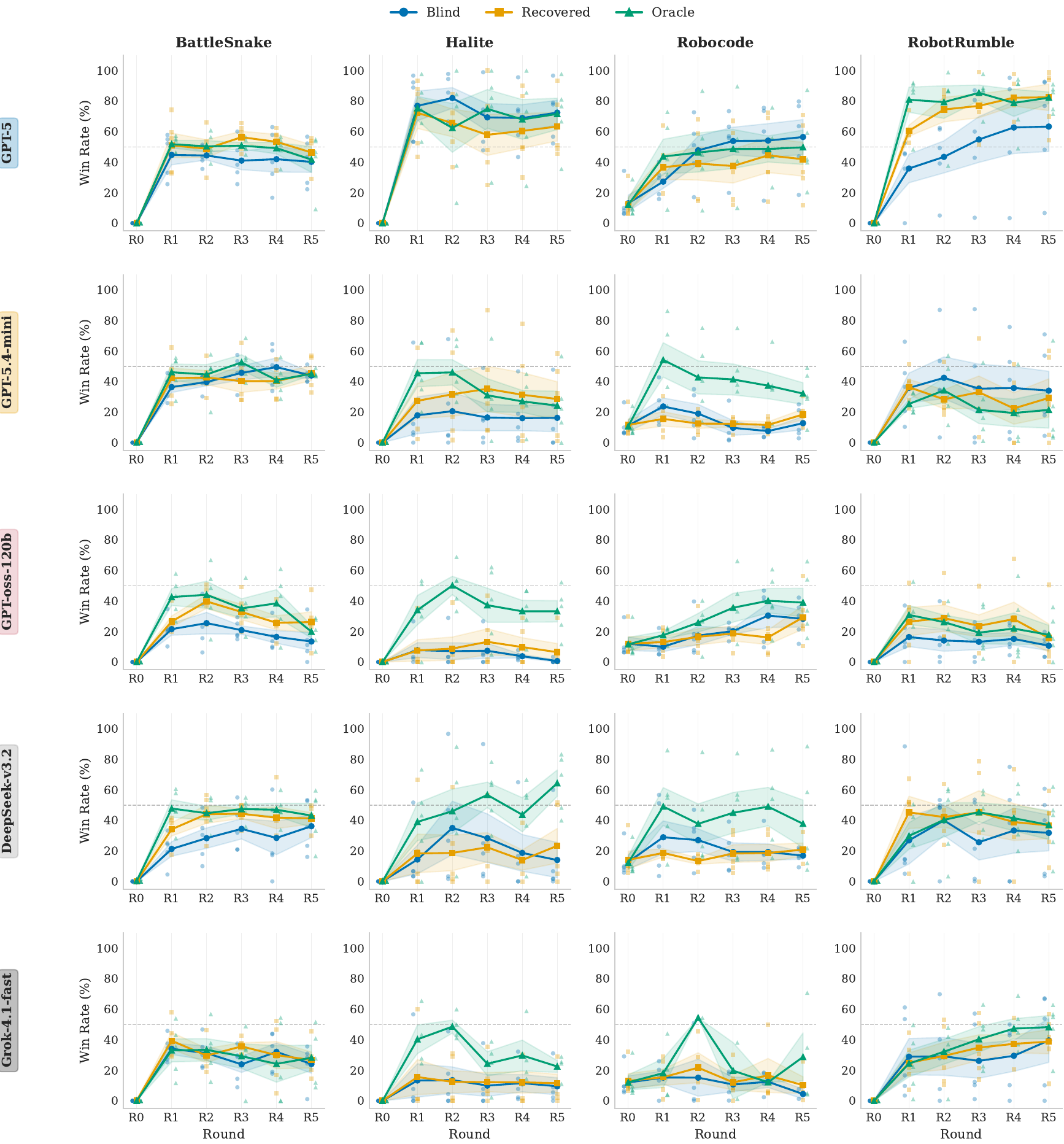}
    \caption{%
        Per-model PvP win-rate trajectories across games and intel conditions (blind, recovered, oracle), ordered from strongest (top) to weakest (bottom).
        Each row is one challenger model; each column is one game.
        Strong models show overlapping conditions or decaying oracle lift; weaker models exhibit persistent or growing oracle lift.
    }
    \label{fig:pvp-per-model}
\end{figure}

Figure~\ref{fig:pvp-per-model} shows per-game challenger win-rate trajectories for each of the five models across all four games and three intel conditions.
Each row corresponds to one challenger model, ordered from strongest (GPT-5) to weakest (Grok-4.1-fast); each column within a row is one game.

\section{Context History Management}
\label{app:history-management}

The inverse-strategy agent operates over multi-round interaction, where each round appends environment observations (evaluation mismatches, probe traces), file reads, and model responses to the conversation history.
Without management, context grows linearly in the number of rounds, risking context-window overflow and inflating inference cost.

We address this with two complementary mechanisms that control context size while preserving the information the agent needs at each editing step: \emph{history compaction}, which truncates and summarises older observations, and a \emph{per-summary cap} that hard-limits any remaining long summaries.

\begin{figure}[h]
    \centering
    \includegraphics[width=0.65\textwidth]{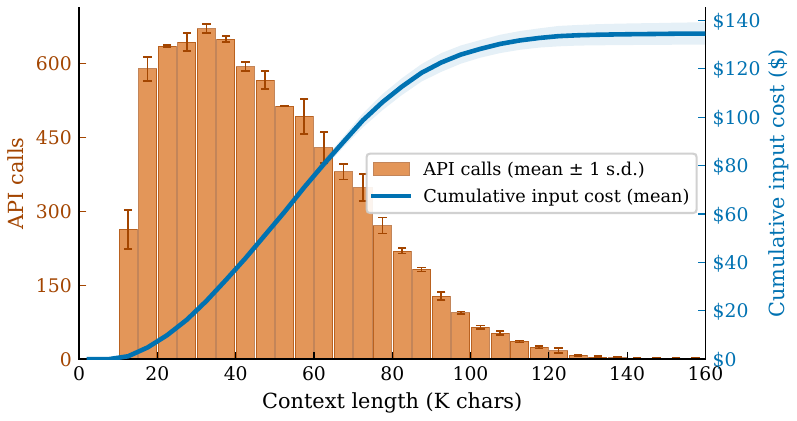}
    \caption{%
        Context-length distribution and cumulative input cost across two independent GPT-5 runs over all targets without compaction (7\,712 and 8\,052 API calls; mean total input cost \$135).
        Bars show the mean call count per 5\,K-char bin (error bars: $\pm$1\,s.d.\ across runs); the blue curve shows cumulative input cost.
        Calls above 60\,K chars account for roughly half the input bill, motivating history compaction.
    }
    \label{fig:context-nocompact}
\end{figure}

\subsection{History Compaction}

History compaction is the primary context-reduction mechanism.
It operates at two levels: an observation-level hard cap applied at \emph{write} time, and a summary-based compaction applied at \emph{query} time on a read-only copy of the conversation.

\paragraph{Observation truncation.}
Before an observation enters the history, each tool output is capped at $L_{\text{out}}$ characters.
Outputs exceeding the limit retain the first and last $L_{\text{out}}/2$ characters; the middle is replaced with an omission marker.
After truncation every observation satisfies:
\begin{equation}
    |o_i| \;\leq\; L_{\text{out}}.
\end{equation}

\paragraph{Summary-based compaction.}
Let $k$ denote the number of recent observations preserved verbatim.
The history is partitioned into a \emph{recent} window (the last $k$ user observations) and a \emph{past} set (all earlier observations).
Each past observation $o_i$ is replaced by a structured summary:
\begin{equation}
    o_i \;\longrightarrow\; \tilde{o}_i = \textsc{Summarize}(o_i).
\end{equation}
The summarizer is rule-based and type-aware: it recognises probe feedback, game traces, source-code reads, and truncated outputs, producing a one-line pointer for each (e.g., \texttt{[Compacted probe 3/5: 42 pairs, 2 sims]} which points to compaction of 42 state-action pairs in 2 simulations).

Because the most recent $k$ observations carry the information the agent is actively working with---the latest mismatch analysis, the last code edit, the most recent probe result---older observations are largely redundant.
Retaining short summaries rather than discarding them entirely prevents the agent from re-exploring strategies it has already tried.

\subsection{Per-Summary Cap}

After compaction, each summarized past observation is further hard-truncated to $L_{\text{past}}$ characters:
\begin{equation}
    \tilde{o}_i \;\longrightarrow\; \hat{o}_i = \textsc{Truncate}(\tilde{o}_i,\; L_{\text{past}}),
    \qquad |\hat{o}_i| \leq L_{\text{past}}.
\end{equation}
This serves as a safety net for cases where the summarizer's output is still large---for example, when a single observation contained interleaved outputs or the original text was so long that even the summary preview is substantial.
Alternatively, it can be used on its own as a lightweight compaction strategy when full summarization is not needed.

\definecolor{savinggreen}{HTML}{2d8a4e}
\definecolor{costred}{HTML}{c0392b}
\definecolor{neutral}{HTML}{666666}

\subsection{History Compaction Ablation}
\label{sec:compaction-ablation}

To isolate the effect of history compaction, we run a controlled ablation on \textsc{BattleSnake} and \textsc{Halite} with GPT-5, comparing three conditions: no compaction ($k{=}\varnothing$), aggressive compaction ($k{=}1$), and moderate compaction ($k{=}3$).
Each condition uses 3 targets $\times$ 2 seeds $=$ 6 runs over 5 rounds.

Figure~\ref{fig:context-nocompact} shows why compaction matters: without it, context lengths reach 150\,K chars with a heavy right tail, and calls above 60\,K chars account for roughly half the input cost.
Figure~\ref{fig:context-compaction} shows the effect of compaction on the context-length distribution: both $k{=}3$ and $k{=}1$ shift the mass leftward, cutting median context from 55\,K to 30\,K and 27\,K chars respectively.

\begin{figure}[t]
    \centering
    \includegraphics[width=0.75\textwidth]{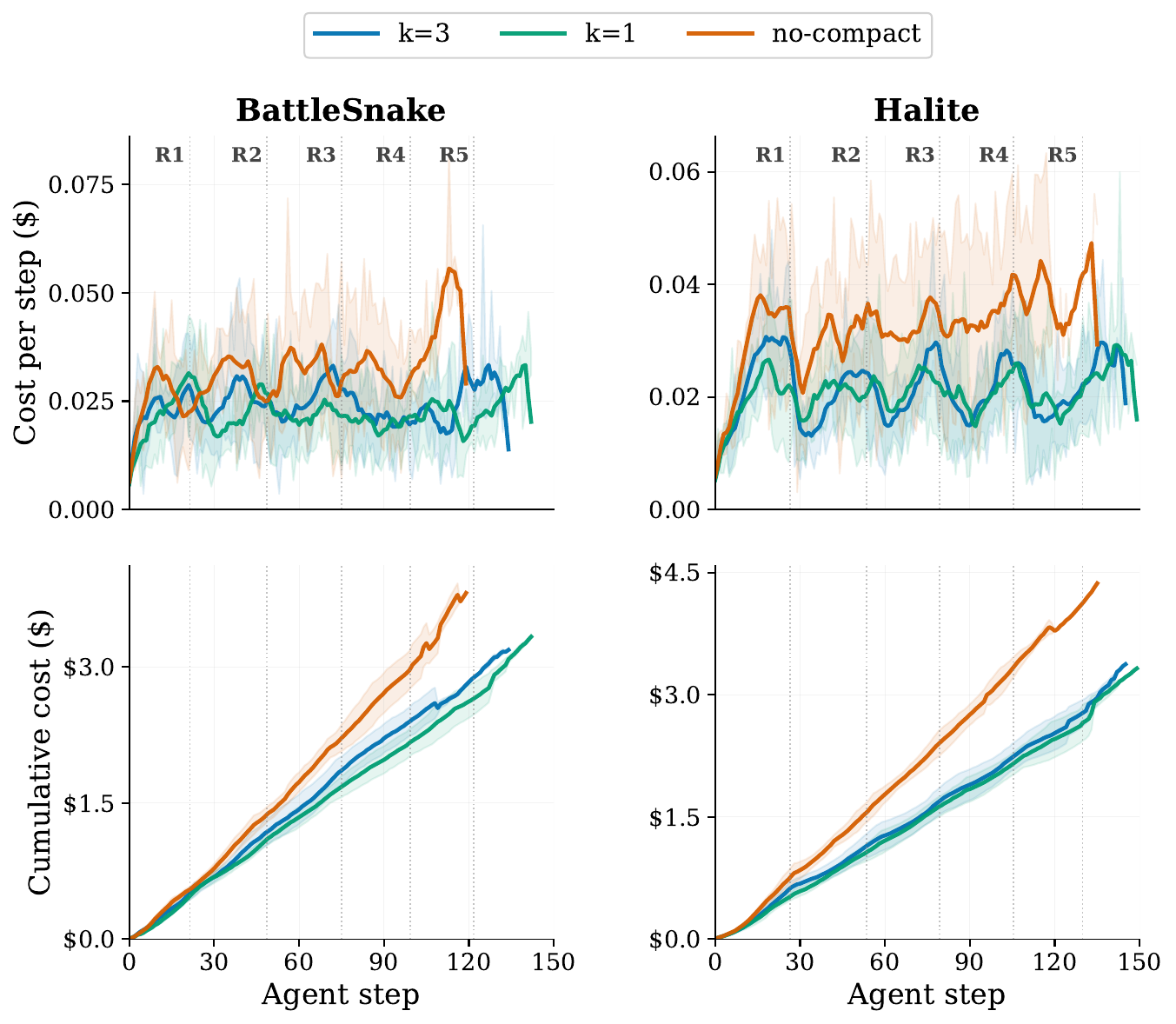}
    \caption{%
        History compaction ablation (GPT-5, 3 targets $\times$ 2 seeds per condition).
        \textbf{Top:} per-step API cost; without compaction (vermillion), cost per call rises as context grows, while both $k{=}1$ and $k{=}3$ keep it flat.
        \textbf{Bottom:} cumulative cost over agent steps.
    }
    \label{fig:compaction-cost}
\end{figure}

\begin{figure}[t]
    \centering
    \includegraphics[width=0.65\textwidth]{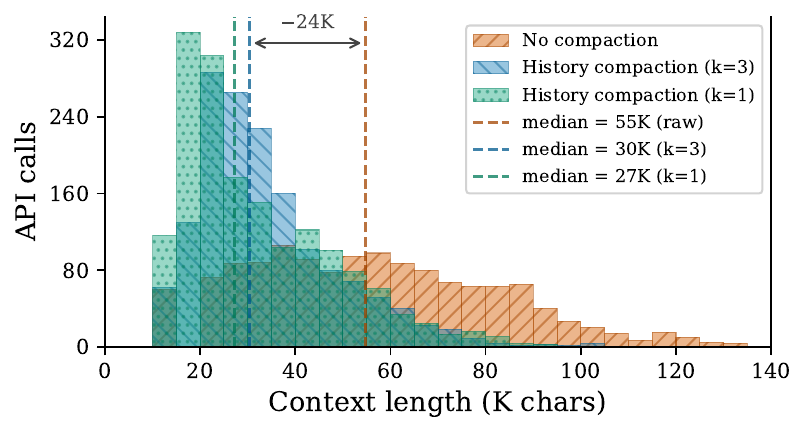}
    \caption{%
        Effect of history compaction on context length (same ablation as Figure~\ref{fig:compaction-cost}).
        Compaction shifts the distribution leftward: median context drops from 55\,K chars (no compaction) to 30\,K ($k{=}3$) and 27\,K ($k{=}1$).
    }
    \label{fig:context-compaction}
\end{figure}

\paragraph{Ablation cost.}
Figure~\ref{fig:compaction-cost} shows the per-step cost trajectory (top row) and cumulative cost (bottom row) for both games.
Without compaction, per-step cost rises steadily as the context grows; both compaction settings keep it flat.
On \textsc{BattleSnake}, $k{=}3$ reduces total cost by 14\% (\$2.87 vs \$3.35) and $k{=}1$ by 11\% (\$2.99);
on \textsc{Halite}, savings reach 20--22\% (\$2.89 and \$2.83 vs \$3.61).
Note that $k{=}1$ is \emph{more} expensive than $k{=}3$ on \textsc{BattleSnake} despite a lower per-step cost: with less context retained, the agent takes more steps per round to reach the same editing decisions, and the additional steps offset the per-call savings.
This effect is more pronounced on \textsc{BattleSnake}, where the agent's editing loop is more sensitive to the amount of recent context available.

\paragraph{Strategy recovery.}
Table~\ref{tab:compaction-summary} summarises cost and behavioral distance across conditions.
Both compaction settings reduce total cost by 11--22\% while producing comparable or better final-round distances.
The $\Delta$ columns show the change relative to the no-compact baseline: negative $\Delta_{\text{cost}}$ indicates savings; negative $\Delta_{d}$ indicates equal or improved strategy recovery.
All distance differences fall well within one standard deviation of the baseline, confirming that summarizing old observations does not degrade performance.
\begin{table}[h]
    \centering
    \caption{History compaction ablation summary (GPT-5, 3 targets $\times$ 2 seeds per condition). $\Delta$ columns show change relative to no-compact. For cost, $\downarrow$ is better; for distance, $\downarrow$ is better.}
    \label{tab:compaction-summary}
    \small
    \begin{tabular}{ll rr rr rr}
        \toprule
        & & \multicolumn{2}{c}{Total cost (\$)} & \multicolumn{2}{c}{R5 distance} & \multicolumn{2}{c}{Best distance} \\
        \cmidrule(lr){3-4} \cmidrule(lr){5-6} \cmidrule(lr){7-8}
        Game & Condition & mean{\scriptsize$\pm$std} & $\Delta$\% & mean{\scriptsize$\pm$std} & $\Delta$ & mean{\scriptsize$\pm$std} & $\Delta$ \\
        \midrule
        \multirow{3}{*}{\textsc{BSnake}}
        & $k{=}3$ & 2.87{\scriptsize$\pm$0.38} & \textcolor{savinggreen}{$-$14\%} & .371{\scriptsize$\pm$.319} & \textcolor{neutral}{+.051} & .272{\scriptsize$\pm$.237} & \textcolor{savinggreen}{$-$.008} \\
        & $k{=}1$ & 2.99{\scriptsize$\pm$0.37} & \textcolor{savinggreen}{$-$11\%} & .456{\scriptsize$\pm$.258} & \textcolor{neutral}{+.136} & .275{\scriptsize$\pm$.239} & \textcolor{savinggreen}{$-$.005} \\
        & no-compact & 3.35{\scriptsize$\pm$0.48} & --- & .320{\scriptsize$\pm$.250} & --- & .280{\scriptsize$\pm$.235} & --- \\
        \midrule
        \multirow{3}{*}{\textsc{Halite}}
        & $k{=}3$ & 2.89{\scriptsize$\pm$0.39} & \textcolor{savinggreen}{$-$20\%} & .549{\scriptsize$\pm$.152} & \textcolor{savinggreen}{$-$.041} & .384{\scriptsize$\pm$.181} & \textcolor{savinggreen}{$-$.086} \\
        & $k{=}1$ & 2.83{\scriptsize$\pm$0.39} & \textcolor{savinggreen}{$-$22\%} & .421{\scriptsize$\pm$.263} & \textcolor{savinggreen}{$-$.169} & .380{\scriptsize$\pm$.204} & \textcolor{savinggreen}{$-$.090} \\
        & no-compact & 3.61{\scriptsize$\pm$0.63} & --- & .590{\scriptsize$\pm$.252} & --- & .470{\scriptsize$\pm$.180} & --- \\
        \bottomrule
    \end{tabular}
\end{table}

\section{Noise-decomposition Methodology}
\label{app:noise-decomposition}

\subsection{Setup}
\label{sec:noise-setup}

We run all 15~targets per arena across the five \benchname{} arenas through Gemma~4~31B with five distinct values of the \verb|tournament.seed| config field (42, 100, 200, 300, 400). The target's behavior is fixed across runs (\verb|fixed_target: true|), and the 20~opponents the target plays in each round~$k$ are picked once by \verb|random.Random(strategy_pool.seed + target_idx)| with \verb|strategy_pool.seed| held at 42 across the five runs, so \textbf{the opponent set per (target, round) is identical across all five runs}. The differences between runs come from (a)~game-engine internal RNG (Halite map generation, BattleSnake food spawns), (b)~Python's global \verb|random.shuffle| in the arena code, and (c)~LLM API non-determinism (no \verb|seed| is passed to OpenRouter). Each (target, run) tuple produces six rounds (round~0~$=$ basis evaluation of the starter code; rounds~1--5~$=$ the agent's edits). Within each round the learner's code is evaluated against the 20~opponent-game traces to produce an action distance per game (\verb|per_simulation|) and a round-level mean (\verb|distance_history[r]|). 75~tournaments~$\times$~6~rounds~$\times$~20~simulations~$=$~9{,}000 sim-level evaluations per arena cell.

\subsection{Variance components}
\label{sec:noise-variance-components}

\begin{figure}[t]
  \centering
  \includegraphics[width=\linewidth]{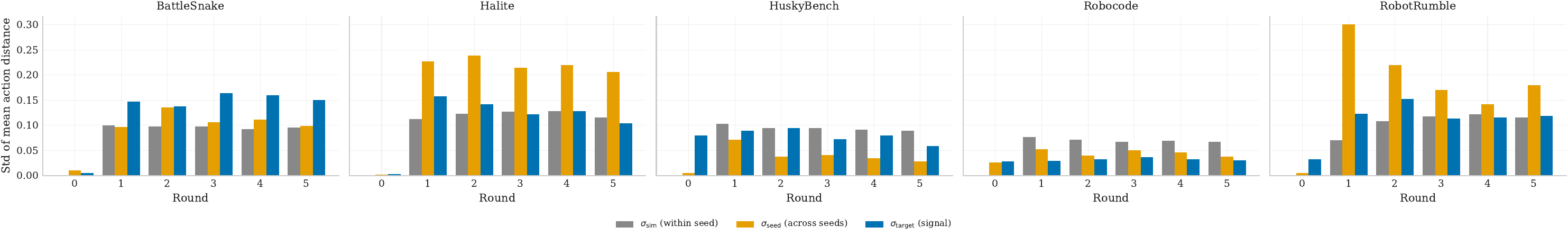}
  \caption{Within-Gemma variance components at each round, grouped side-by-side (not stacked, since standard deviations do not add). $\sigma_{\rm sim}$ is the within-run, across-opponent spread; $\sigma_{\rm seed}$ is the across-run process-level noise at fixed target and opponent set; $\sigma_{\rm target}$ is the across-target signal. Round~0 has near-zero $\sigma_{\rm seed}$ and $\sigma_{\rm target}$ because the starter code is identical across runs and basis difficulty is uniform within an arena.}
  \label{fig:noise-decomposition}
\end{figure}

Let $d_{tsri}$ be the action distance of target~$t$'s code at round~$r$ against opponent~$i$ in run~$s$. We report standard deviations $\sigma$ (\texttt{ddof}~$=1$) rather than variances so units stay in the metric's native scale (action distance $\in [0,1]$).
\begin{itemize}
  \item \textbf{Within-run ($\sigma_{\rm sim}$).} $\sigma_{\rm sim}^{(t,s,r)} = \mathrm{sd}_i \, d_{tsri}$, then averaged over $(t,s)$ at the requested round. Holds the learner's code, the target's behavior, and the run fixed; only the opponent identity and per-sim RNG vary. Captures the irreducible spread of one tournament round across its opponent set.
  \item \textbf{Across-run ($\sigma_{\rm seed}$).} $\sigma_{\rm seed}^{(t,r)} = \mathrm{sd}_s \, \bar d_{tsr}$, averaged over targets. Holds the target and the opponent set fixed; varies only the run. Captures the residual process-level non-determinism after averaging over the 20~opponents.
  \item \textbf{Across-target ($\sigma_{\rm target}$).} $\sigma_{\rm target}^{(r)} = \mathrm{sd}_t \, \bar d_{t \cdot r}$ where the inner mean is over runs. The signal: how much targets actually differ in difficulty.
\end{itemize}
We report all three at round~5 (the final round) in Table~\ref{tab:within-model-noise}; Figure~\ref{fig:noise-decomposition} shows them as grouped bars per round, not stacked, since standard deviations do not add. This split parallels the across-task / within-task decomposition recently formalised for one-shot QA evaluation \citep{wang2026noise}.

\begin{table}[t]
\centering
\caption{Within-Gemma noise decomposition at round~5 across the five arenas (15~targets, 5~runs each). $\sigma_{\rm sim}$, $\sigma_{\rm seed}$, and $\sigma_{\rm target}$ are standard deviations of the round-5 mean action distance with the inner aggregation specified in \S\ref{sec:noise-variance-components}. ICC and $\bar\rho$ are computed on the (target $\times$ run) matrix of normalised recovery score (\S\ref{sec:noise-headline}). Bootstrap CI uses 2{,}000 hierarchical resamples.}
\label{tab:within-model-noise}
\begin{tabular}{lrrrrrr}
\toprule
Arena & $\sigma_{\rm sim}$ & $\sigma_{\rm seed}$ & $\sigma_{\rm target}$ & ICC & $\bar\rho$ & recovery (95\% CI) \\
\midrule
BattleSnake & 0.096 & 0.098 & 0.150 & 0.81 & 0.88 & 66.4\% [54.7, 78.0] \\
Halite      & 0.115 & 0.206 & 0.104 & 0.07 & 0.14 & 63.1\% [55.6, 69.3] \\
Poker  & 0.089 & 0.028 & 0.059 & 0.85 & 0.84 & 68.5\% [55.5, 80.4] \\
RoboCode    & 0.067 & 0.038 & 0.030 & 0.41 & 0.32 & 33.7\% [27.9, 39.8] \\
RobotRumble & 0.116 & 0.179 & 0.118 & 0.19 & 0.23 & 30.5\% [22.3, 38.6] \\
\bottomrule
\end{tabular}
\end{table}

Two arenas (BattleSnake, Poker) have $\sigma_{\rm target}$ larger than both noise sources and ICC~$> 0.8$; their headline metric is signal-dominated. Three arenas (Halite, RoboCode, RobotRumble) have $\sigma_{\rm seed}$ comparable to or larger than $\sigma_{\rm target}$ and ICC~$\leq 0.41$; scores there move substantially run-to-run, so single-run point estimates should be replaced with run-aggregated means or paired-test framings when comparing models.

\subsection{Headline statistics}
\label{sec:noise-headline}

\begin{figure}[t]
  \centering
  \includegraphics[width=\linewidth]{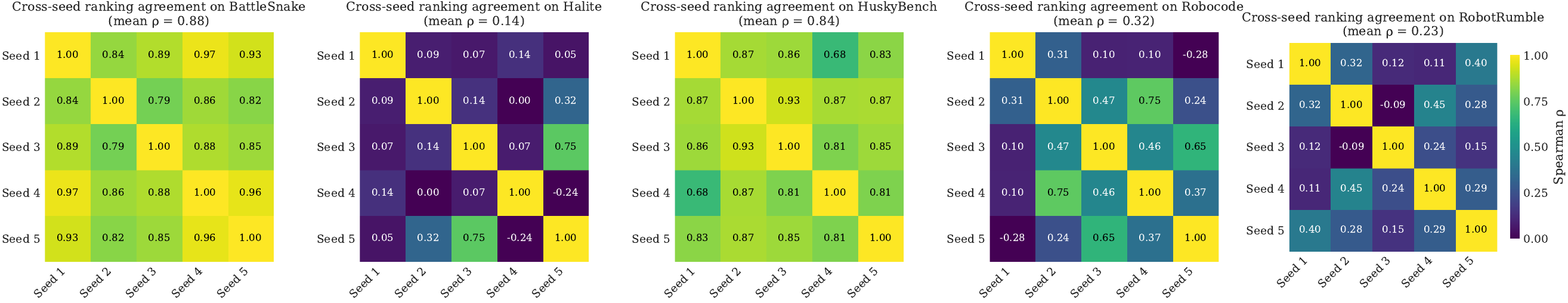}
  \caption{Within-Gemma cross-run Spearman~$\rho$ between per-target normalised-recovery rankings, one panel per arena. Mean $\rho$ in each panel title. BattleSnake and Poker show uniformly high agreement; Halite, RoboCode, and RobotRumble show large run-to-run rank flips.}
  \label{fig:ranking-stability}
\end{figure}

We compute consistency ICC (Shrout--Fleiss two-way random-effects, single-rater) on the (target $\times$ run) matrix of $\mathrm{bri}_{ts}$ values, where
\[
  \mathrm{bri}_{ts} = 100 \cdot \frac{d_{ts0} - \min_{r \ge 1} d_{tsr}}{d_{ts0}}
\]
is the normalised recovery score over the basis distance. Closed form:
\[
  \mathrm{ICC} = \frac{\mathrm{MS}_{BT} - \mathrm{MS}_{E}}{\mathrm{MS}_{BT} + (k-1)\,\mathrm{MS}_{E}},
\]
where $\mathrm{MS}_{BT}$ is the between-target mean square and $\mathrm{MS}_{E}$ is the residual mean square (run main effect excluded; this is the consistency form rather than absolute agreement). ICC near 1 means signal dominates noise; ICC near 0 means the opposite. We additionally report the mean pairwise Spearman~$\rho$ between per-target rankings across runs (full $5 \times 5$ matrix per arena in Figure~\ref{fig:ranking-stability}). The 95\% confidence interval on the mean recovery score is from a hierarchical bootstrap (2{,}000 resamples: targets with replacement, then runs within each chosen target).

\subsection{Cross-model validity}
\label{sec:cross-model}

The within-Gemma analysis above describes how reliably one model is scored. To test whether the benchmark itself assigns consistent difficulties to its targets across evaluators, we apply the same Spearman~$\rho$ and ICC machinery across our 12-model lineup at one run per model. Gemma~4~31B contributes its 5-run mean as a stabilised reference; the other 11 evaluators (DeepSeek~v4 Pro\,/\,Flash, GLM-5.1, GPT-5.5, GPT-5, GPT-5.4-mini, GPT-oss-120b, Gemma~4 26B-A4B, Grok-4.1-fast, Kimi~K2.6, Qwen3.6~35B-A3B) contribute their \verb|seed=42| run.

\begin{table}[t]
\centering
\caption{Cross-model agreement on per-target normalised recovery score at round~5. $N_{\rm models}=12$ (Gemma~4~31B at its 5-run mean; the other 11 at their single seed=42 run). ICC$_{\rm cross}$ is consistency ICC on the (target $\times$ model) matrix; $\bar\rho_{\rm cross}$ is the mean pairwise Spearman~$\rho$ between models on per-target rankings.}
\label{tab:cross-model}
\begin{tabular}{lrrrr}
\toprule
Arena & $N_{\rm models}$ & $N_{\rm targets}$ & ICC$_{\rm cross}$ & $\bar\rho_{\rm cross}$ \\
\midrule
BattleSnake & 12 & 15 & 0.32 & 0.51 \\
Halite      & 12 & 15 & 0.20 & 0.33 \\
Poker  & 12 & 15 & 0.58 & 0.61 \\
RoboCode    & 12 & 15 & 0.07 & 0.09 \\
RobotRumble & 12 & 15 & 0.11 & 0.14 \\
\bottomrule
\end{tabular}
\end{table}

The ordering tracks the within-Gemma reliability picture. Poker is the only arena with strong cross-evaluator agreement (ICC 0.58, $\bar\rho = 0.61$), and BattleSnake is moderate ($0.32$, $0.51$); the original within-Gemma claim that BattleSnake rankings are run-stable does not extend cleanly to evaluator-stability. Halite, RoboCode, and RobotRumble disagree across evaluators on per-target difficulty (ICC $\leq 0.20$), tracking the same high run-to-run noise visible in the within-Gemma sweep rather than reflecting evaluator-specific behavior.

\begin{figure}[t]
  \centering
  \includegraphics[width=\linewidth]{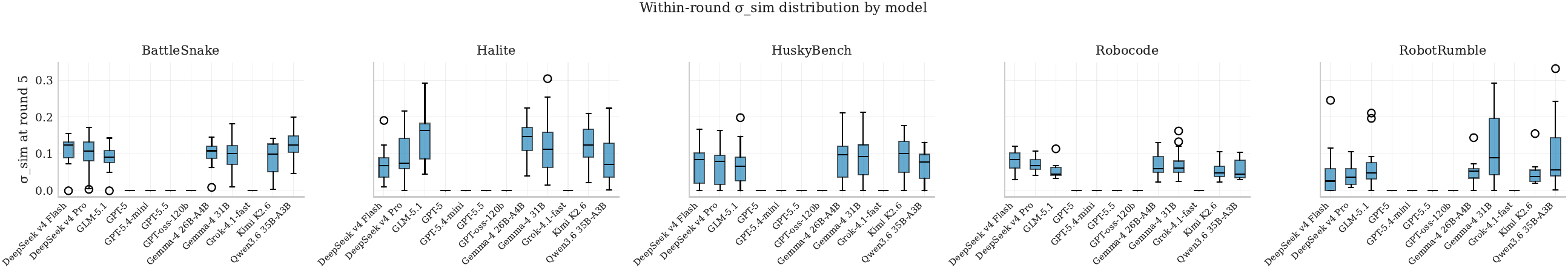}
  \caption{Distribution of $\sigma_{\rm sim}$ at round~5, one boxplot per evaluator, faceted by arena. The within-arena overlap across the 12 evaluators supports treating $\sigma_{\rm sim}$ as an arena-level property rather than a model-specific quantity.}
  \label{fig:sigma-sim-per-model}
\end{figure}

\paragraph{$\sigma_{\rm sim}$ is an arena-level property.} Figure~\ref{fig:sigma-sim-per-model} shows the $\sigma_{\rm sim}$ distribution at round~5 per (model, arena). Within each arena the 12 per-model boxplots overlap heavily, confirming that the irreducible opponent-draw noise depends on the arena's mechanics, not on the evaluator. The within-Gemma $\sigma_{\rm sim}$ values reported above therefore generalise to the other 11 evaluators.

\begin{figure}[t]
  \centering
  \includegraphics[width=0.72\linewidth]{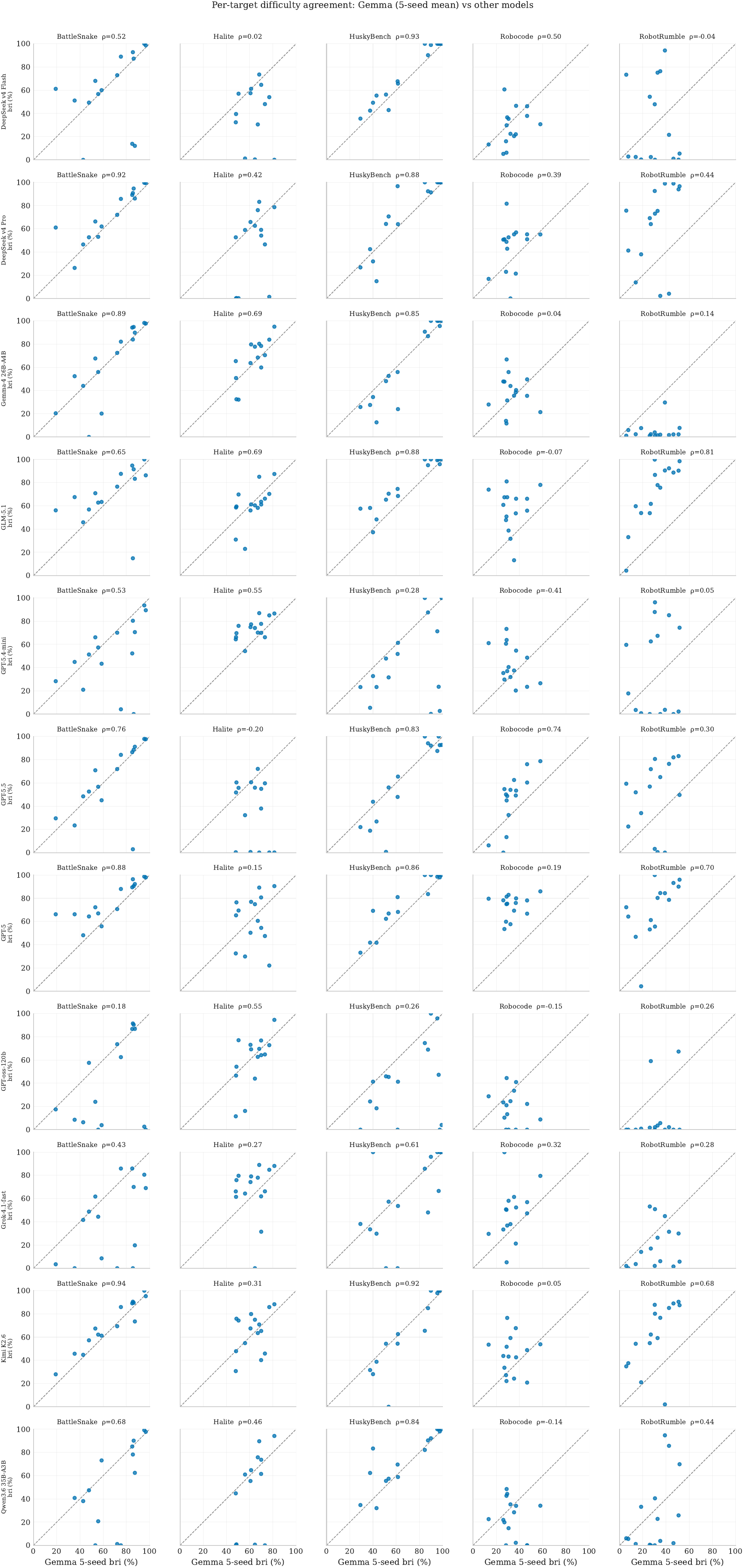}
  \caption{Per-target difficulty agreement: Gemma~4~31B's 5-run mean recovery score (x-axis) versus each other evaluator's single-run recovery (y-axis), one row per other model and one column per arena. Dashed line is the 1:1 reference; Spearman $\rho$ is in each panel title. On Poker most $\rho \geq 0.6$; on RoboCode and RobotRumble many $\rho$ are below 0.3 with sign flips, consistent with the low cross-model ICC in Table~\ref{tab:cross-model}.}
  \label{fig:target-difficulty-agreement}
\end{figure}

\subsection{Limitations}

\textbf{Five runs is a small sample.} Recent reproducibility studies of small reasoning benchmarks argue for $K \approx 30$ runs as a stabilisation point for per-question pass-rate estimates \citep{hochlehnert2025a}; our 5-run budget is well below that, so we treat per-cell tests as descriptive and lean on the aggregate bootstrap and ICC for the headline reliability claims.

textbf{Only one model has multi-run coverage.} The 11 additional evaluators contribute one run each, so the within-arena $\sigma_{\rm seed}$ reported here is Gemma-specific. The $\sigma_{\rm sim}$ consistency check (Figure~\ref{fig:sigma-sim-per-model}) is the closest evidence that the rest of the noise-structure characterisation generalises.

textbf{Outlier sensitivity.} Some runs fail catastrophically: the agent's context window overruns, its code crashes the engine, or its tournament hits the step\,/\,cost budget mid-way. We keep these in the analysis because a model that destroys its own context or submits broken code is exhibiting real (negative) capability. On the Gemma~4~31B sweep, RoboCode has 14\,/\,75 runs ($\approx 19\%$) truncated, Halite has 5~runs with round-5 distance $\geq 0.85$, Poker has 4~truncated runs, and BattleSnake and RobotRumble have none. The headline numbers therefore mix two distinct sources: noise on successful runs, and a binary ``this run failed'' signal. Table~\ref{tab:within-model-noise} should be read as upper bounds on the noise share that survives outlier removal, particularly for RoboCode (truncations) and Halite (high-distance failures).

\section{Cost Analysis}
\label{app:cost-analysis}

\begin{figure}[!h]
    \centering
    \includegraphics[width=0.75\linewidth]{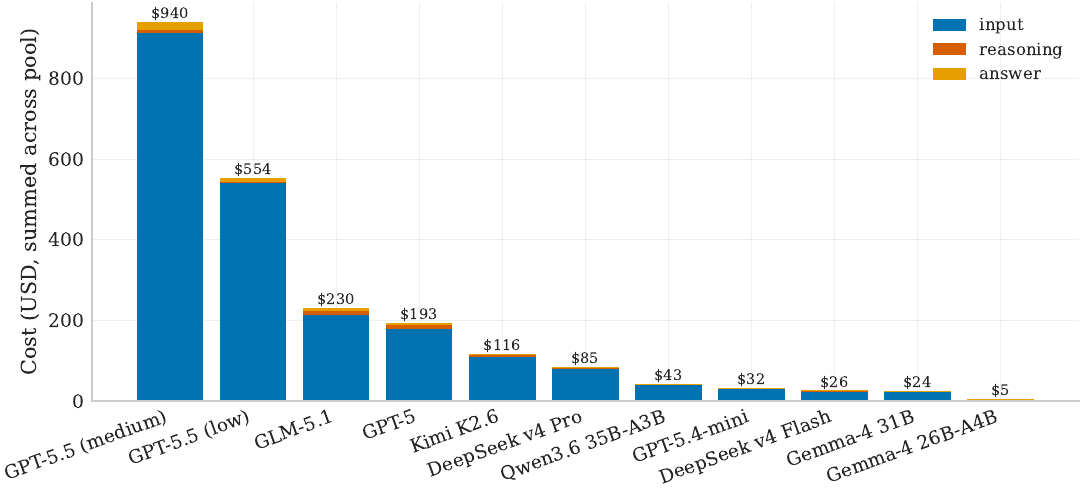}
    \caption{Total API cost breakdown per model across all 75 runs.
    Stacked segments show prompt, reasoning, and answer token charges.}
    \label{fig:cost-breakdown}
\end{figure}

\begin{figure}[!h]
    \centering
    \includegraphics[width=0.75\linewidth]{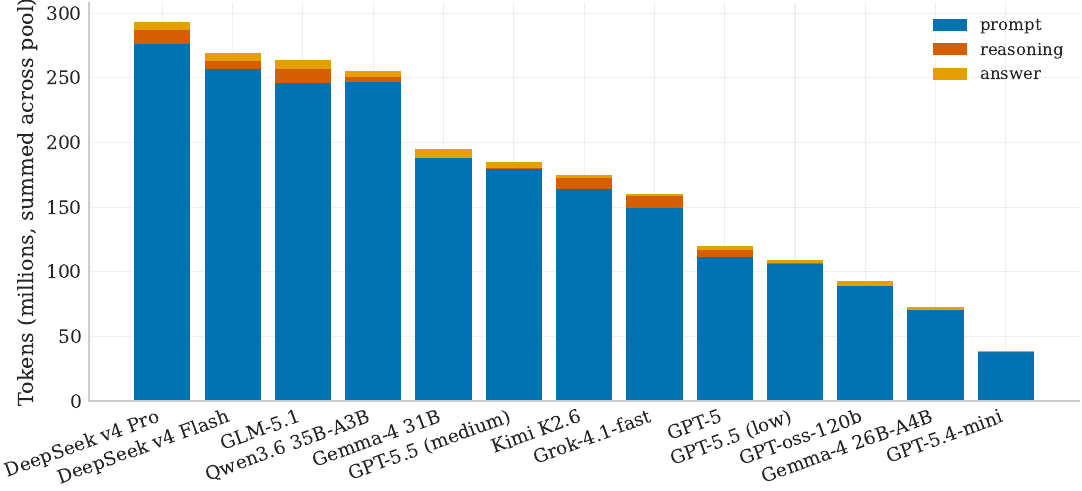}
    \caption{Total token usage per model (millions).
    Stacked segments show prompt, reasoning, and answer tokens.
    DeepSeek v4 Pro is the most token-intensive model; GPT-5.4-mini the most frugal.}
    \label{fig:token-usage}
\end{figure}

All models are evaluated on the same pool of 75 target strategies (15 per game $\times$ 5 games).
Figures~\ref{fig:cost-breakdown} and~\ref{fig:token-usage} visualise the cost and token-usage breakdowns.

\paragraph{Inference cost.}
Figure~\ref{fig:cost-breakdown} breaks down the total API cost per model into prompt (input), reasoning, and answer (output) token charges.
GPT-5.5 (medium) is the most expensive configuration at \$940 for the full pool, driven by its high reasoning-effort setting and per-token pricing.
GPT-5.5 (low) costs \$554, while the Codex variants (low and high) cost only \$5 and \$0 respectively due to Codex's bundled pricing.
GPT-5.4-mini remains highly cost-efficient at \$43.
DeepSeek v4 Pro and Kimi K2.6 occupy a middle ground at \$116 and \$193 respectively.

\paragraph{Token usage.}

Figure~\ref{fig:token-usage} shows raw token consumption summed over all 75 runs.
DeepSeek v4 Pro consumes the most tokens overall ({$\sim$}293\,M), reflecting both long prompts from its multi-turn context window and substantial reasoning overhead.
GPT-5.5 (Codex high) and GPT-5.5 (Codex low) use the fewest tokens ({$\sim$}38--72\,M), benefiting from Codex's more efficient scaffold.
GPT-5.5 (medium) uses {$\sim$}194\,M tokens despite being the most expensive, indicating that cost is driven by per-token pricing rather than volume alone.
GPT-5.4-mini remains token-frugal at {$\sim$}72\,M.

\end{document}